\pdfoutput=1

\documentclass[11pt]{article}

\usepackage[]{acl}

\usepackage{times}
\usepackage{latexsym}

\usepackage[T1]{fontenc}

\usepackage[utf8]{inputenc}

\usepackage{microtype}

\usepackage{inconsolata}

\usepackage{amsmath,amsfonts,bm}
\usepackage{graphicx}
\usepackage{multirow}
\usepackage{makecell}
\usepackage{array}
\usepackage[para]{threeparttable}
\usepackage{booktabs}
\usepackage{tabularx}
\usepackage{xspace}
\usepackage{caption}
\usepackage{subcaption}
\usepackage{longtable}
\usepackage{hyperref}
\usepackage{bbm}
\usepackage{bm}
\usepackage{enumitem}
\usepackage{algorithm}
\usepackage{algpseudocode}
\usepackage{geometry}
\usepackage{comment}
\usepackage[colorinlistoftodos]{todonotes}
\usepackage{tablefootnote}
\usepackage{color}
\usepackage{tcolorbox}
\usepackage{tikz}
\usepackage{tabularray}
\usepackage[normalem]{ulem}

\usepackage[T1]{fontenc}
\usepackage[utf8]{inputenc}
\usepackage{listings}
\usepackage{xcolor}
\usepackage{xspace}

\newcommand{\ConAtomic}[1]{C_{A_{#1}}}
\newcommand{\ConGold}[1]{C_{G_{#1}}}
\newcommand{\PredAtomic}[1]{P_{A_{#1}}}

\definecolor{main}{HTML}{5989cf}    
\definecolor{sub}{HTML}{cde4ff}    

\tcbset{
    sharp corners,
    colback = white,
    before skip = 0.2cm,   
    after skip = 0.5cm     
}    

\newtcolorbox{boxC}{
    colback = sub, 
    boxrule = 0pt 
}

\title{How Well Do Large Language Models Truly Ground?}

\author{Hyunji Lee$^1$$^{*}$ \quad Se June Joo$^1$$^{*}$ \quad Chaeeun Kim$^1$$^{\dag}$ \quad Joel Jang$^2$ \\ \textbf{Doyoung Kim$^1$} \quad \textbf{Kyoung-Woon On$^3$} \quad \textbf{Minjoon Seo$^1$} \\
 \\
$^1$KAIST AI \quad $^2$University of Washington \quad $^3$Kakao Brain  \\
\texttt{\{hyunji.amy.lee, sejune, minjoon\}@kaist.ac.kr} 
}

\begin{document}
\maketitle
\def\thefootnote{*}\footnotetext{Denotes equal contribution}
\def\thefootnote{\dag}\footnotetext{Work done during internship at KAIST AI}
\def\thefootnote{\arabic{footnote}}

\begin{abstract}
To reduce issues like hallucinations and lack of control in Large Language Models (LLMs), a common method is to generate responses by grounding on external contexts given as input, known as knowledge-augmented models. However, previous research often narrowly defines ``grounding'' as just having the correct answer, which does not ensure the reliability of the entire response. To overcome this, we propose a stricter definition of grounding: a model is \textit{truly} grounded if it (1) fully utilizes the necessary knowledge from the provided context, and (2) stays within the limits of that knowledge. We introduce a new dataset and a grounding metric to evaluate model capability under the definition. We perform experiments across 25 LLMs of different sizes and training methods and provide insights into factors that influence grounding performance. Our findings contribute to a better understanding of how to improve grounding capabilities and suggest an area of improvement toward more reliable and controllable LLM applications\footnote{Our code and data are available at \href{https://github.com/kaistAI/How-Well-Do-LLMs-Truly-Ground}{https://github.com/kaistAI/How-Well-Do-LLMs-Truly-Ground}}.
\end{abstract}

\section{Introduction}
\label{sec:intro}
Large Language Models (LLMs) have shown superior performance on various tasks by leveraging the extensive world knowledge embedded in their parameters. However, these models often produce hallucinations~\citep{Bender2021OnTD, Du2023ImprovingFA}, lack controllability~\citep{Dathathri2019PlugAP, Zhang2022ASO}, and have trouble integrating knowledge that changes over time~\citep{Lin2021TruthfulQAMH, Wang2021CanGP}. Additionally, they may not contain specialized knowledge unique to certain entities, such as company-specific terminology, or private information not contained in the training data. Although it is technically possible to inject new knowledge by further training LLMs on a specific corpus, this approach is generally inefficient and not practical in many scenarios~\citep{Mallen2022WhenNT, Panda2023DifferentiallyPI, Tang2023PrivacyPreservingIL}.
To address these issues, various systems~\footnote{\url{https://www.bing.com/new}, \url{https://www.perplexity.ai/}, \url{https://openai.com/blog/chatgpt-plugins}} and work~\citep{Gao2023EnablingLL, He2022RethinkingWR, Xu2023SearchintheChainTA, yao2022react} have explored methods where such dynamic, specialized, or private contexts provided by users or general world knowledge contexts retrieved from a large corpus (retrieval-augmented models) are provided to LLMs as additional inputs. 

While previous work has shown enhanced performance by allowing LLMs to ground their outputs on external contexts compared to solely relying on the LLM's inherent knowledge~\citep{andrew2007scalable, BehnamGhader2022CanRL, Mallen2022WhenNT}, whether the model \textit{well-grounds} to the contexts is usually measured by simply checking whether the generated response contains the answer~\citep{Liu2023LostIT, Mallen2022WhenNT, Lewis2020RetrievalAugmentedGF} or evaluating over NLI model to see whether the knowledge from given context correlates with generated response~\citep{Gao2023EnablingLL, Asai2023SelfRAGLT}. However, in some cases, this may not be sufficient and it may be more important to ensure that the \textit{entire} generated response is \textit{truly} grounded on the given external contexts.

\begin{figure*}[ht!]
    \centering
    \begin{minipage}[b]{0.98\textwidth}
    \includegraphics[width=\textwidth]{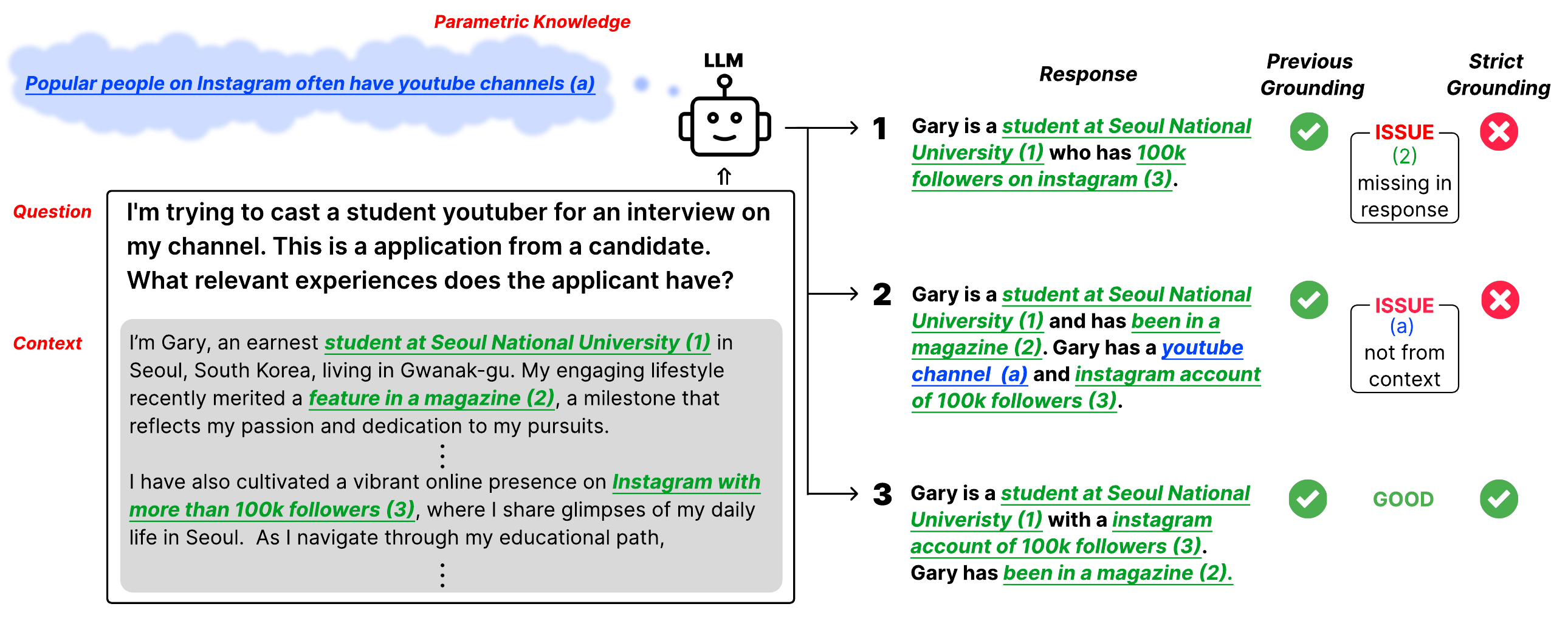}
    \caption{ \fontsize{6.5}{10}\footnotesize 
    An example scenario of a company's HR team using LLM to question upon candidate's resume which is given as input context.  
    The previous definition of grounding would consider responses 1 and 2 as well grounded due to their high relevancy with the question and input context. However, as our definition considers all knowledge in a fine-grained manner, we consider \textit{only} response 3 as well-grounded.
    Response 1 misses key resume detail (2) which makes the candidate underrated. Response 2 introduces knowledge (a) that is not from the given context but from the model's parametric knowledge, inaccurately overrates the candidate, and unfairly influences comparison with others.}
    \label{fig:main}
    \end{minipage}
\end{figure*}

For example, let's consider the scenario in Figure~\ref{fig:main}, where a company's HR team is utilizing an LLM to question the qualifications of candidates by providing their resumes as external contexts and prompting the LLM to provide an answer to questions about the candidates based on their resumes. Response 1 omits essential information about the candidate and Response 2 contains misinformation about the candidate due to generating knowledge contained in its parameters; both cases do not truly represent the candidate's qualifications. It either harms the applicant by missing important information or makes the applicant overly qualified, disadvantaging other applicants.

In this study, we introduce a strict definition of grounding: a model is \textit{truly} grounding on given contexts when it (1) uses all essential knowledge from the contexts and (2) strictly adheres to their scope in response generation without hallucinated information\footnote{In this paper, the term grounding refers to what is defined here as truly grounding.}. 
To quantify this definition, we introduce an automatic grounding metric that extends upon \citet{Min2023FActScoreFA} for fine-grained evaluation. 
Furthermore, we curate a new dataset incorporating crucial factors influencing LLMs' response (i.e., entity popularity, context length), to understand their impact on LLM responses. Lastly, we present a revised version of the dataset that modifies factual knowledge in external contexts to identify the knowledge sources in responses.

We conduct experiments across 25 LLMs of different sizes and training methods to explore which model attributes significantly contribute to grounding ability and identify some important factors. 
\begin{itemize}[noitemsep,leftmargin=1.8em]
    \item Training methods like Instruction Tuning or RLHF have a more pronounced impact on grounding performance than model size.
    \item High answer accuracy, commonly used to assess how well a model incorporates context in previous works, does not ensure high grounding performance.
    \item Instruction-tuned models show high degradation when additional relevant contexts are added as input.
    \item When given multiple contexts, performance degradation is more influenced by how distracting these contexts are, rather than by their length.
\end{itemize}

\section{Related Works}

\paragraph{Question Answering}

Machine Reading Comprehension and Open Domain Question Answering provide a question and context to a model, which then answers the question using the given context. The answers are usually short phrases or entities. LongformQA shares similarities, as it also uses contextual information to answer questions, but its answers are longer and focus on how well the model refers to the input context and generates factual responses.
Such datasets, while encompassing questions and contexts, are inadequate to measure the model's grounding ability under our definition; they lack annotation of which knowledge from the external context is necessary (gold) to answer the query and are hard to verify the source of knowledge in generated response (whether it is from a given context or model parameter). Furthermore, since most datasets were created before the emergence of modern LLMs, they're unsuitable for understanding the diverse characteristics of these models. Therefore, to evaluate a model's grounding ability under our defined criteria, we created a new dataset.

\paragraph{Generating Response with External Knowledge}

Recent research efforts have focused on incorporating external knowledge during the generation process to overcome issues such as hallucination, increase controllability, and incorporate dynamic knowledge. 
It incorporates either by inputting it directly~\citep{Lewis2020RetrievalAugmentedGF, liu2023evaluating, Shi2023REPLUGRB}, using APIs in a multi-step manner~\citep{yao2022react, Xu2023SearchintheChainTA}, or by employing various tools~\citep{Schick2023ToolformerLM, Yang2023GPT4ToolsTL}. Although the objective of adding external knowledge is for the model's response to be intrinsically tied to the given knowledge, previous work naively evaluates and analyzes the ability.
With such a naive definition, users find it difficult to ensure that the entire generated response is truly grounded in the given context; the model may hallucinate or miss important knowledge even though the overall response corresponds well to the external context.
Thereby, in this work, we introduce a strict definition of grounding and share the importance of checking the entire response in a fine-grained manner.

\paragraph{Definition of Grounding}
The concept of "grounding" pervades several areas that interface with natural language. 
In robotics, grounding bridges the chasm between abstract directives and actionable robot commands, as highlighted by numerous studies~\cite{Ahn2022DoAI,Huang2023GroundedDG,Kollar2010TowardUN,Kollar2010GroundingVO,Tellex2011UnderstandingNL,Mees2022GroundingLW, faille-etal-2021-entity-based, moon-etal-2019-opendialkg, brabant2023kgconv, gc, theory}. 
In the domain of vision and video, grounding predominantly involves associating image regions with their pertinent linguistic descriptors~\cite{Zhu2022SeqTRAS,Deng2021TransVGEV,Li2022InvariantGF,Liu2022EntityEnhancedAR}. 
In NLP, grounding frequently denotes finding the relevant textual knowledge to a given input from knowledge sources such as a set of documents, knowledge graphs, or input context~\citep{Chandu2021GroundingI,Weller2023AccordingT,Mallen2022WhenNT}; information retrieval task. In this work, we focus on bridging the definition with when input context is the knowledge source.

\section{Grounding}

In this paper, we define that the model grounds well more strictly and share a dataset and metric to measure performance under the definition. In Section~\ref{sec3: def}, we define the grounding ability and share its importance with various use cases. In Section~\ref{sec3: data}, we share details of how we construct the dataset, and in Section~\ref{sec3: Metric}, we formulate an automatic metric to measure the grounding ability. 
\subsection{Definition \& Usage} \label{sec3: def}
Prior research~\citep{Liu2023LostIT, He2022RethinkingWR, Mallen2022WhenNT, Weller2023AccordingT} defines that a model is well-grounded when it generates responses relevant to the query while utilizing the given contexts.
When given a set of external contexts $\mathcal{C}$, a set of answers $\mathcal{A}$, and generated response $P$, the previous definition often defines it well-grounded if $\forall a \in \mathcal{A}, \; a \in P$ or $\exists c \in \mathcal{C} : \text{NLI}(P, c) = 1$. The former calculates whether the generated response contains all answers and the latter measures whether any context entails the generated response.
However, as in Figure~\ref{fig:main}, we can see that such a definition of grounding poses limitations in that it cannot capture whether the generated response misses relevant knowledge from a given context or whether it hallucinates.
In this work, to overcome the limitation, we formally define a stricter definition of a model's grounding performance, which evaluates the entire generated response in a fine-grained manner.

We define that a model \textit{truly} grounds on provided external context when (1) it utilizes \textit{all} necessary knowledge in the context, and (2) it does \textit{not} incorporate other knowledge apart from the contexts, such as that stored in the model parameters.
Here, we see the ``atomic facts'' (short sentences conveying one piece of information) as the knowledge unit. As a sentence contains multiple knowledge, we disassemble\footnote{Following \citet{Min2023FActScoreFA}, we use InstructGPT (text-davinci-002) on decomposing context into atomic facts, where it has shown a high correlation with humans. Examples of atomic facts are in Appendix~\ref{appendix: atomic facts}.} a single sentence into multiple atomic facts for a fine-grained evaluation~\citep{Min2023FActScoreFA, Liu2022RevisitingTG, Kamoi2023WiCERE}. 
For instance, ``Napoleon is a French general'' decomposes into two atomic facts (``Napoleon is French.'' and ``Napoleon is a general.''). 

In other words, when given a set of necessary atomic facts (gold atomic facts) $\mathcal{C}_{G}$ from the set of external contexts $\mathcal{C}$ and a set of atomic facts $\mathcal{P}_{A}$ from the generated response $P$, we define that the model is \textit{truly} grounded when: \\
\begin{enumerate}[noitemsep,leftmargin=1.8em] 
\vspace{-1em}
\item $\forall k \in \mathcal{C}_{G}, k \in P$ \\
\vspace{-1em}
\item $\forall k \in \mathcal{P}_{A}, \exists c \in \mathcal{C} \text{ such that } k \in c$ \\ 
\vspace{-1em}
\end{enumerate}

\begin{figure*}[ht!]
    \centering
    \begin{minipage}[b]{1.0\textwidth}
        \centering
    \includegraphics[width=1.0\textwidth]{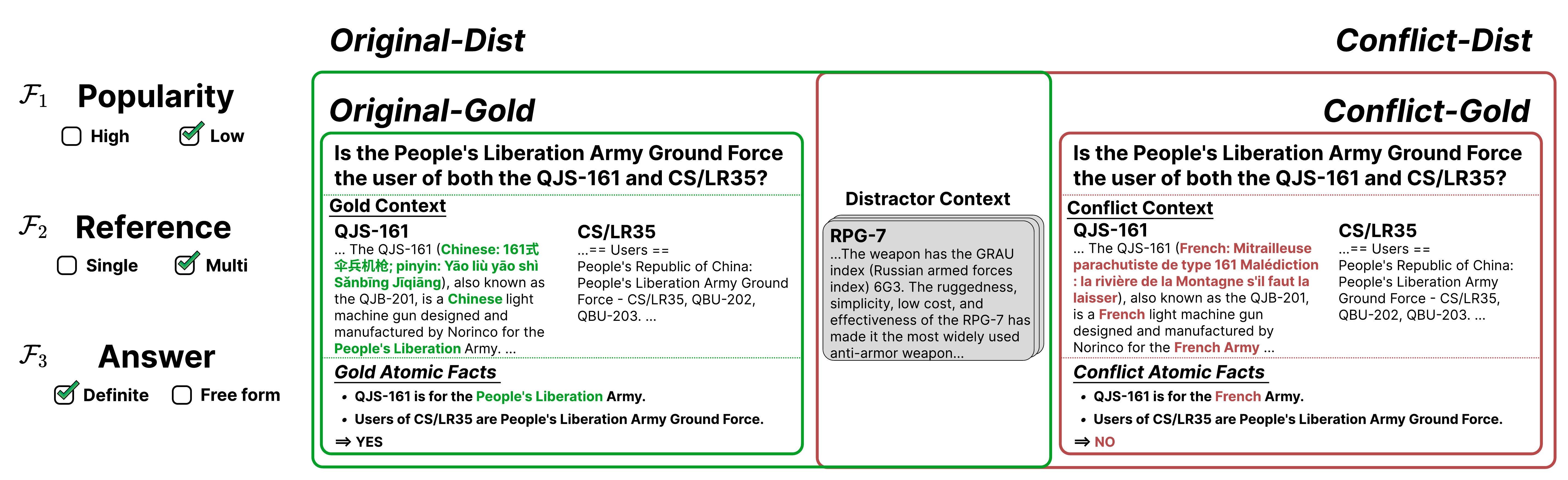}
    \caption{\fontsize{6.5}{10}\footnotesize Four versions of our dataset: \textit{Original-Gold}, \textit{Original-Dist}, \textit{Conflict-Gold}, and \textit{Conflict-Dist}. Conflict-* contains modified gold contexts (conflict context) by human annotators. *-Dist differs from *-Gold in that it contains distractor contexts. The left part of the figure shows three key factors we considered when constructing our dataset.}
    \label{fig:data_ex}
    \end{minipage}
    \vspace{-2em}
\end{figure*}

Models that demonstrate strong grounding capabilities as per our definition are highly valued in various use cases. 
It can be used in developing personalized chatbot services. 
By grounding contexts with personal information, it adeptly uses it to generate responses. When new information is provided by the user, it can be seamlessly integrated into the input context for future interactions.
Also, when a company wants to add advertisement by promoting a certain product; by providing the model with the necessary context, it can be guided to generate responses that favorably mention the product. 
Moreover, models with a strong grounding ability allow users to trust the responses generated without the need to verify for inaccuracies or omissions, effectively addressing the issue of hallucinations.

\subsection{Dataset Construction} \label{sec3: data}

We construct a new evaluation dataset specifically designed to measure a model's grounding ability due to limitations of existing datasets; they lack annotation of which knowledge from the provided context is necessary, hard to verify the source of knowledge (whether the knowledge is from a given context or its parameter), and most do not consider key variables known to influence LLM performance as they were constructed before the advent of modern LLM. 

As in Figure~\ref{fig:data_ex}, our dataset comprises four versions: \textit{Original-Gold}, \textit{Original-Dist}, \textit{Conflict-Gold}, and \textit{Conflict-Dist}. 
The differentiation lies in two main aspects: (1) The nature of the input context, which is either an unaltered Wikipedia content (\textit{Original-*}) or a modified, conflicting version (\textit{Conflict-*}) to determine whether the model's response is from its internal knowledge or by grounding on external knowledge. (2) The inclusion of distractor contexts: \textit{* -Gold} versions contain only ``gold contexts'' that directly answer the query, whereas \textit{*-Dist} versions also include distractor contexts, which are relevant but not gold.

Furthermore, we integrate three key factors (left of Figure~\ref{fig:data_ex}) known to bring qualitative differences in model responses for a more comprehensive analysis: [$\mathcal{F}_1$] Popularity of context topics ~\citep{Mallen2022WhenNT, Kandpal2022LargeLM}, [$\mathcal{F}_2$] Number of required documents to answer the query~\citep{BehnamGhader2022CanRL, Press2022MeasuringAN, Cfka2022BlackboxLM}, and [$\mathcal{F}_3$] Required response format (definite answer or free-form answer)~\citep{McCoy2021HowMD, Tuckute2022SentSpaceLB}.

Our dataset construction is mainly divided into five steps. Details of data construction including human annotators, inter-labeler agreement, data distribution of the factors, data examples, and more are in Appendix~\ref{app: overall_data}.
\paragraph{Step 1: Context Selection}
In our first step, we select sets of input contexts ($\mathcal{C}$) considering $\mathcal{F}1$ and $\mathcal{F}_2$. Wikipedia documents were used for context, considering their comprehensive meta-information pertinent to these aspects. For $\mathcal{F}1$, following \citet{Mallen2022WhenNT}, we utilize document pageviews, and for $\mathcal{F}_2$, we construct a document set sampled from the intersection between the popularity list and the hyperlinked document.
\paragraph{Step 2: Instance Generation \& Classification}
Based on the document sets from Step 1,  we use GPT-3.5\footnote{gpt-3.5-turbo-0301} to generate 10 candidate pairs of question and answer. We classify the candidate pairs by $\mathcal{F}_2$ and $\mathcal{F}_3$, and select a single query with the highest quality from each class. 
Note that the generated answer was replaced by the annotators. 
\paragraph{Step 3: Gold Atomic Fact Selection}
To evaluate grounding performance, we decompose context sets $C \in \mathcal{C}$ into atomic facts $\{\ConAtomic{1}, \cdots, \ConAtomic{k}\}$. 
From multiple atomic facts, we annotate \textit{gold} atomic facts, $\ConGold{i}$. Gold atomic facts are the atomic facts within the provided context that are essential to answer the given question ($\{\ConGold{1}, \cdots, \ConGold{m}\} \subseteq \{\ConAtomic{1}, \cdots, \ConAtomic{k}\}$). We now get 480 complete instances that we call \textit{Original-Gold} $(Q,A,\mathcal{C},\mathcal{C}_{G})$.
\paragraph{Step 4: Modify Context}
Given an instance from \textit{Original-Gold}, annotators are instructed to revise well-known and key knowledge to answer the question in the input context. 
This step results in \textit{Conflict-Gold} $(Q,A',\mathcal{C}',\mathcal{C}_{G}')$, a modified, conflicting version. 
\paragraph{Step 5: Add Distractor Contexts}
To analyze the impact when additional knowledge apart from the gold ones is added to the input context, we sample distractor contexts, contexts with high similarity but not directly related to an answer, with contriever~\cite{izacard2022unsupervised}, a dense retriever pretrained through contrastive learning, and include them in the input context (\textit{Original-Dist} when added to original gold contexts and \textit{Revised-Dist} when added to revised gold contexts).
 
\subsection{Metric} \label{sec3: Metric}
We evaluate model performance in two aspects: grounding performance and answer accuracy.

\begin{table*}[ht!]
\centering
\fontsize{6.5}{10}\selectfont
    \begin{tabular}{ccccccccccccccc}
    \toprule
        Size & \multicolumn{5}{c}{7B} & \multicolumn{3}{c}{13B} & \multicolumn{1}{c}{40B} & \multicolumn{2}{c}{70B} & \multicolumn{2}{c}{UNK}  \\
            \cmidrule(lr){2-6} \cmidrule(lr){7-9} \cmidrule(lr){11-12} \cmidrule(lr){13-14}
         $M_{pred}$  & Llama2-C & Vicuna & \textsc{T\"ulu2}  & Mistral-I & Zephyr  & Llama2-C & Vicuna & \textsc{T\"ulu2}  & Falcon-I  & Llama2-C & \textsc{T\"ulu2} & GPT & GPT-I\\
    \midrule
    Original-Gold & 51.6 & 50.0 & 58.6  &60.3 & 54.7  & 55.9 & {61.4} & \underline{61.9} & 42.4  & 56.9 & \underline{61.9} & 61.0 & \textbf{65.7} \\
    Original-Dist& 45.1 & 45.0 & 54.9  & 54.9 & 53.7 & 35.8 & {56.5} & 55.3 & 36.3 & 55.8 & \underline{56.7} & 56.8 & \textbf{56.9} \\
    Conflict-Gold & 46.0 & 48.0 & 54.9  & 59.8 & 52.4 & 53.4 & {57.5} & 57.7 & 40.1  & 56.3 & \underline{\textbf{62.4}} & 59.0 & {60.3} \\
    Conflict-Dist & 40.4 & 39.8 & 47.9 & 54.3& 52.4 & 46.5 & \underline{55.0} & 50.4 & 32.6 & 54.4 & 54.9 & \textbf{56.1} & 54.5 \\
    \bottomrule
    \end{tabular}
\caption
     {\fontsize{6.5}{10}\footnotesize Grounding performance of twelve different models. For each setting, the best of all in \textbf{bold} and the best of open-sourced models in \underline{underline}.} 
\label{table: f1_score}
\end{table*}

\paragraph{Grounding Performance}

We present an automatic metric to measure whether the model grounds well under the definition in Section~\ref{sec3: def}. 
We evaluate the presence of knowledge (whether an atomic fact exists in context) by using an evaluation model $M_{eval}$, as the same facts can be conveyed in different ways. 
On selecting $M_{eval}$ we use the one with the highest correlation with humans.
We test over five models: GPT-4~\citep{OpenAI2023GPT4TR}, Llama-2-70b-chat~\citep{Touvron2023Llama2O}, TRUE (T5-11B finetuned on various NLI datasets)~\citep{Honovich2022TRUERF}, bi-encoder model (MiniLM finetuned on 1B training pairs), and cross-encoder model (MiniLM finetuned on MSMARCO)~\citep{Wang2020MiniLMDS}. 
Surprisingly, the cross-encoder model\footnote{cross-encoder/ms-marco-MiniLM-L-12-v2 from Sentence Transformers~\citep{reimers-2019-sentence-bert}} shows the highest correlation with human (84.1), outperforming GPT-4 (78.7). It also closely matches the correlation between humans (88.6)
Thereby, we utilize the cross-encoder model as $M_{eval}$.

We define grounding performance as the \textbf{F1 score} of precision and recall calculated as:
$\text{precision} = \sum_{i=1}^{k} M_{eval} (\PredAtomic{i}, C)$ and $\text{recall} = \sum_{i=1}^{m} M_{eval} (\ConGold{i}, P)$
\noindent where $M_{eval}(a, B)$ returns 1 when knowledge of $a$ exists in $B$ and 0 elsewise. 
Details of models, performance, and the process of human evaluation are in Appendix~\ref{appendix: human_eval}.

\paragraph{Answer Accuracy}
This is a widely used metric to naively measure the model's grounding ability in previous works~\citep{Mallen2022WhenNT, Borgeaud2021ImprovingLM}; it measures if the answer is present within the generated response\footnote{We only measure the metric to queries with definite answers.}.

\section{Experiments}

\begin{table}[h]
\centering
\fontsize{6.5}{10}\selectfont
    \begin{tabular}{cccccc}
    \toprule
         & Base & DPO & RLHF & Inst. & Size \\
    \midrule
    Llama2 & Llama2 & x & x & x & [13] \\
    Llama2-C & Llama2&  x & o & o & [7, 13, 70] \\
    Vicuna & Llama2&  x & x & o & [7, 13, 33] \\
    \textsc{T\"ulu1} & Llama1 &  x & x & o & [7, 13, 30, 65] \\
    \textsc{T\"ulu2} & Llama2 &  x & x & o  & [7, 13, 70] \\
    \textsc{T\"ulu2}-D & Llama2&  o & x & o& [7, 13, 70] \\
    Falcon & Falcon &  x & x & x & [40, 180]\\
    Falcon-I & Falcon&  x & x & o & [40, 180]\\
    Mistral-I & Mistral & x & x & o  & [7]\\
    Zephyr & Mistral & o & x & o & [7] \\
    \bottomrule
    \end{tabular}
\caption
     {\fontsize{6.5}{10}\footnotesize Abstract of open-sourced LLMs we experiment over. The size column shows various sizes of the model we experimented over. The base column shows the pretrained model each model is finetuned on. The rest of the columns show different training methods; Inst. is instruction-tuned, DPO is Direct Preference Optimization, and RLHF is Reinforcement Learning from Human Feedback. } 
     \vspace{-1em}
\label{table: models}
\end{table}

We experiment with 25 LLMs of various sizes and training methods (Instruction-tuning, RLHF, DPO).
From the results, we share interesting findings of how different factors of LLMs and different characteristics of input context lead to their grounding ability. 
Section~\ref{sec4: models} shows brief details of the models we evaluate. 
Section~\ref{sec4: results} shows how different factors of LLMs lead to their grounding ability and interesting findings. Details of the input format, generation configurations, and others are in Appendix~\ref{app: inference}.

\subsection{Models} \label{sec4: models}

We experiment with two proprietary LLMs: GPT-3.5 (GPT) and GPT-3.5-instruct (GPT-I)\footnote{Specific model names for each model were gpt-3.5-turbo-0301 and gpt-3.5-turbo-instruct. Further detail can be found at \url{https://platform.openai.com/docs/models}}. The latter, GPT-instruct\footnote{After this point, we shorten GPT-3.5 to ''GPT''}, is a further finetuned version of GPT, primarily for following instructions. 
Table~\ref{table: models} shows details of open-sourced LLMs we experiment over: Llama2~\citep{Touvron2023Llama2O}, Llama2-chat (Llama2-C), Vicuna, \textsc{T\"ulu1}~\citep{Wang2023HowFC}, \textsc{T\"ulu2}~\citep{Ivison2023CamelsIA}, \textsc{T\"ulu2} with DPO (\textsc{T\"ulu2}-D), Mistral-Instruct (Mistral-I)~\citep{Jiang2023Mistral7}, Zephyr~\citep{Tunstall2023ZephyrDD}, Falcon~\citep{Penedo2023TheRD}, and Falcon-Instruct (Falcon-I). All checkpoints are provided from huggingface~\citep{Wolf2019HuggingFacesTS}.

\subsection{Results} \label{sec4: results}

\paragraph{Overall performance}

Table~\ref{table: f1_score} shows the overall grounding performance of various models over four different dataset versions\footnote{Details of each dataset scenarios in Section~\ref{sec3: data}}. Due to limited space, the results of all models in four dataset versions are in Appendix~\ref{app: revised_gold}. GPT-I shows the highest performance for original datasets (\textit{Original-Gold} and \textit{Original-Dist}), and \textsc{T\"ulu2}-70B shows the highest performance among open-sourced models, similar performance with GPT. 
Performance of \textit{Conflict-Gold} consistently shows lower performance than \textit{Original-Gold} (average of 4.7 drops), which we hypothesize is due to conflict between parametric space and external knowledge. 
The performance also consistently degrades with distractor contexts added: an average of 10.7 drops for \textit{Original-Dist} from \textit{Original-Gold} and an average of 10.0 drops for \textit{Conflict-Dist} from \textit{Conflict-Gold}. The drop is higher than when given conflicting knowledge, which highlights the LLM's tendency to deviate from the primary context when presented with extraneous information and the importance of providing only the gold contexts for high grounding performance. 
When comparing the different model sizes of the same model (i.e., \textsc{T\"ulu2} and Llama-C), the grounding performance of all four dataset versions tends to steadily increase. The improvement rate by a larger model tends to be stronger as the dataset is difficult; \textit{Conflict-Dist} is considered more difficult over \textit{Original-Gold} as it contains more knowledge in input context and contains conflict knowledge with its parametric space. 
When comparing the performance of precision and recall, a common trend across all models is a superior performance in precision over recall (Appendix~\ref{app: prec_and_recall}). This suggests a challenge in utilizing all necessary knowledge when generating a response and it tends to utilize only a partial of them. 

\begin{figure*}[ht!]
    \centering
    \begin{minipage}[b]{0.42\textwidth}
        \centering
    \includegraphics[width=1.0\textwidth]{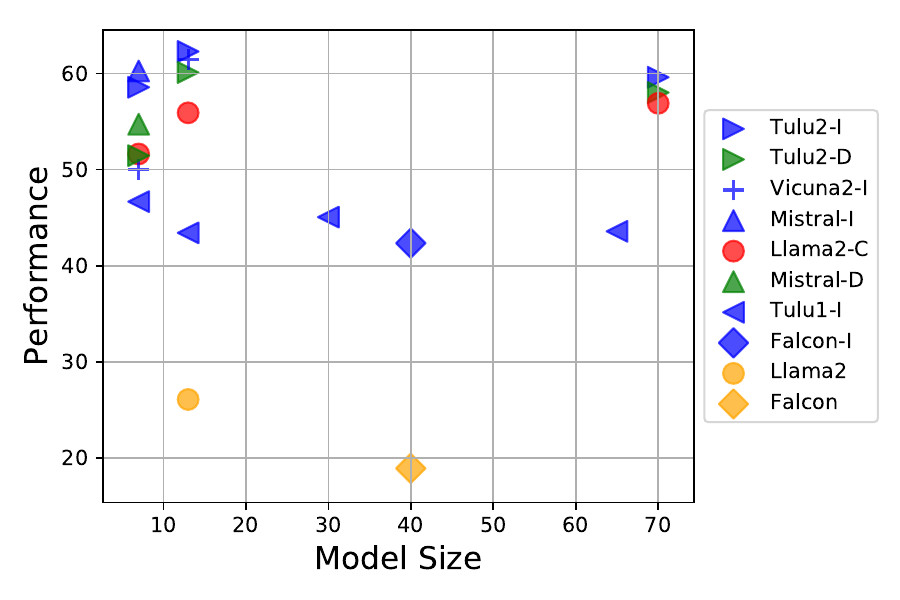}
    \label{fig:overall.model_size}
    \centering 
    (a)
    \end{minipage}
    \begin{minipage}[b]{0.42\textwidth}
        \centering
    \includegraphics[width=1.0\textwidth]{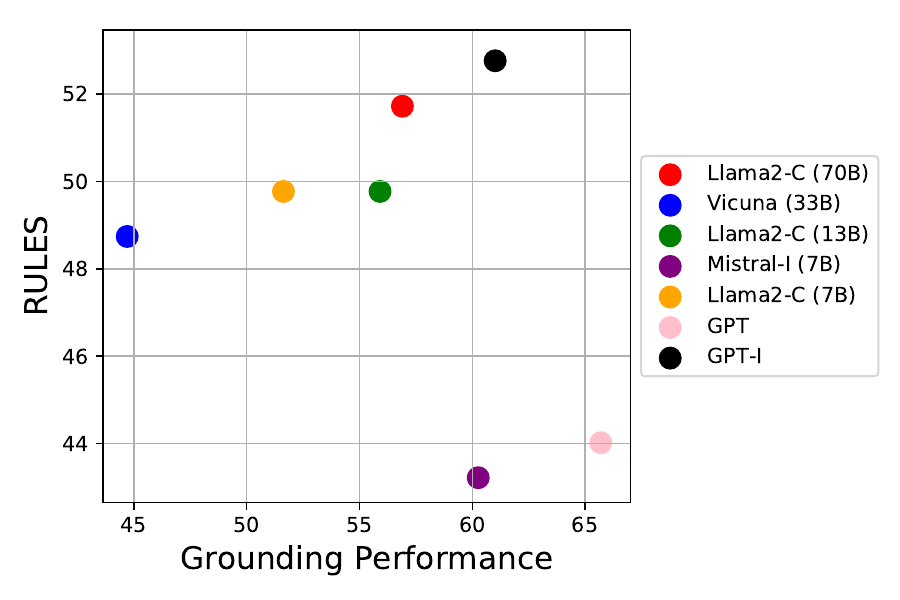}
    \centering
    (b)
    \label{fig:overall.RULES}
    \end{minipage}
    \begin{minipage}[b]{0.8\textwidth}
        \centering
    \includegraphics[width=\textwidth]{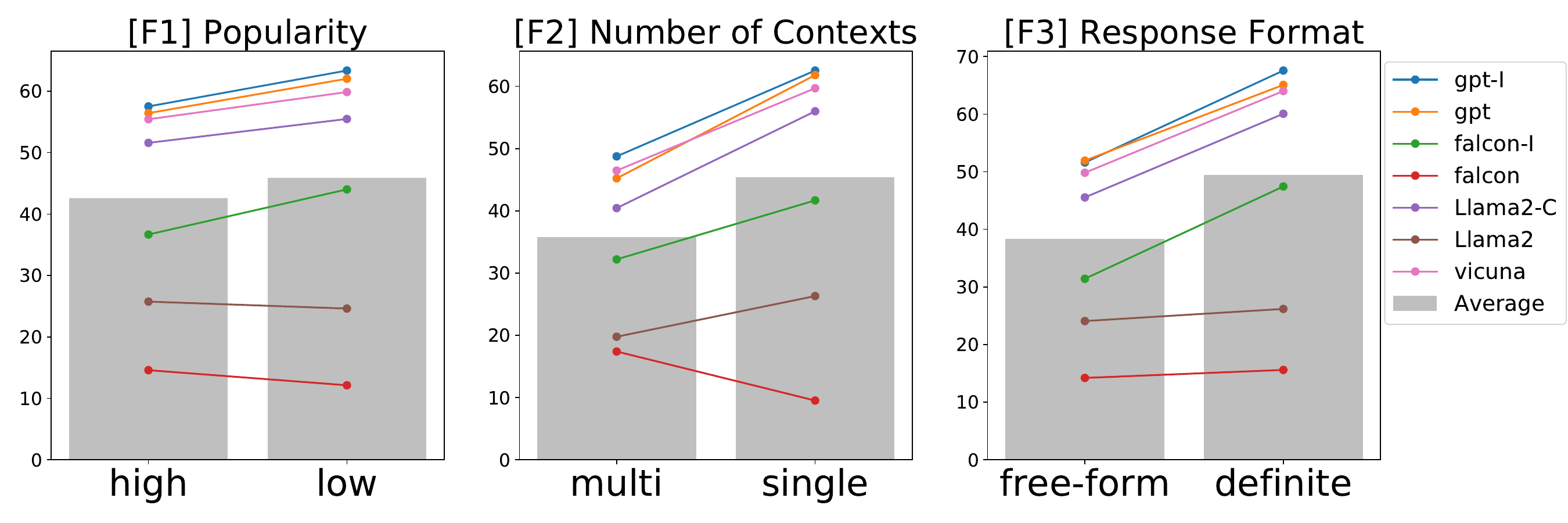}
    \centering 
    (c)
    \label{fig:overall.RULES}
    \end{minipage}
    \caption{\fontsize{6.5}{10}\footnotesize \textbf{(a)} shows grounding performance for each model size in \textit{Original-Gold}. The performance tends to depend more heavily on how the model was tuned rather than the model size. \textbf{(b)} shows RULES performance and grounding performance. There is a weak correlation between instruction-following ability and grounding performance. \textbf{(c)} shows details of grounding performance by the characteristics of queries and contexts in \textit{Original-Gold}. Llama2 and Vicuna are 13B, Falcon is 40B model.}
    \label{fig: overall}
\end{figure*}

\paragraph{Training method shows stronger effect than model size in grounding performance}

Figure~\ref{fig: overall} \textbf{(a)} shows that model size tends to show a small effect on the grounding performance of \textit{Original-Gold}, but how the model was tuned tends to show a stronger effect; for high grounding performance, instruction tuning seems to be the most important factor. 
To determine if grounding performance is strongly dependent on instruction-following ability, we see the correlation between grounding performance with performance on RULES benchmark~\citep{Mu2023CanLF}, a benchmark to determine how well it follows the given rule. Figure~\ref{fig: overall} \textbf{(b)} shows that there is weak correlation between the two scores. This suggests that grounding performance does not appear to be strongly reliant on the capacity to adhere to instructions. We could see a similar trend with MMLU benchmark~\citep{Hendrycks2020MeasuringMM} in Appendix~\ref{app: MMLU}.

\paragraph{Grounding performance by different query and context characteristics}

Figure~\ref{fig: overall} \textbf{(c)} displays the detailed analysis of each model's grounding performance of \textit{Original-Gold}, over the three factors described in Section~\ref{sec3: data}. A consistent trend emerges across all models.
For $\mathcal{F}_1$, the model generally outperforms when provided with less common contexts (low), compared to when provided with more prevalent contexts (high). This resonates with \citet{Mallen2022WhenNT}, underlining a model's propensity to lean on provided data when faced with less familiar content. 
For $\mathcal{F}_2$, queries demanding reasoning across multiple contexts (multi) show lower grounding performance than those confined to a single context (single). The grounding challenges likely arise from the extended context length in multiple scenarios and the added reasoning complexity to extract all relevant atomic facts. 
Lastly, for $\mathcal{F}_3$, questions with predetermined answers (definite) tend to achieve better grounding than open-ended answers (free-form). This divergence largely stems from recall metrics as free-form instances contain more necessary knowledge (gold atomic facts) compared to definite instances, it is more difficult to find all.
We could see that the trend holds for all four dataset settings in Appendix~\ref{app: revised_gold}.

\paragraph{High answer accuracy does not ensure high grounding performance}
Answer accuracy is a common metric used for measuring the grounding ability of a model. However, though there is a correlation between grounding performance (Table~\ref{table: f1_score}) and answer accuracy (Table~\ref{table: ans_acc}), high answer accuracy does not ensure high grounding performance as grounding performance in the same range of answer accuracy highly diverges.
For example, the answer accuracy of Llama2-13b-chat (84.79) and Llama2-13b (81.56) only show a marginal difference of 3.23 compared to the difference of 29.82 (55.91, 26.09) in grounding performance. This discrepancy is attributed to Llama2-13b’s tendency to generate lengthy responses with relevant information drawn not only from the provided context but also its internal parameters, leading to lower grounding scores despite high answer accuracy. 

\paragraph{Smaller models tend to show a higher reduction rate by DPO training}

\begin{table}[ht!]
\centering
\fontsize{6.5}{10}\selectfont
    \begin{tabular}{c|ccc|ccc}
    \toprule
    & \textsc{T\"ulu2} & + DPO & \textbf{deg.rate} (\%) & \textsc{T\"ulu2} & + DPO & \textbf{deg.rate} (\%) \\
    \midrule
    & \multicolumn{3}{c}{\textit{Original-Gold}} & \multicolumn{3}{c}{\textit{Revised-Gold}} \\
    \midrule
    7B & 56.2 & 51.5 & 8.5 & 54.9 & 51.4 & 6.4  \\
    13B & 62.3 & 60.1 & 3.5 & 61.9 & 58.0 & 6.3 \\
    70B & 59.6 & 58.0 & 2.7 & 59.9 & 58.1 & 3.1 \\
    \midrule
    & \multicolumn{3}{c}{\textit{Original-Dist}} & \multicolumn{3}{c}{\textit{Revised-Dist}} \\
    \midrule    
    7B & 54.9 & 45.3 & 17.6 & 47.9 & 41.4 & 13.5  \\
    13B & 55.3 & 54.0 & 2.3& 50.4 & 54.2 & -7.5 \\
    70B & 53.4 & 55.4 & -3.7 & 52.4 & 55.1 & -5.1 \\
    \bottomrule
    \end{tabular}
\caption
     {\fontsize{6.5}{10}\footnotesize Grounding performance of \textsc{T\"ulu} and those trained with DPO (+DPO). \textbf{deg.rate} column shows the degradation rate from \textsc{T\"ulu} to those trained with DPO.} 
\label{table: dpo}
\end{table}
Table~\ref{table: dpo} shows the degradation rate from \textsc{T\"ulu2} to those trained with DPO. Smaller models tend to show a higher degradation rate in grounding performance by DPO training. 
The degradation rate tends to come from its verbosity, aligning with the findings from \citet{Ivison2023CamelsIA}.
Moreover, the results of Zephyr, a 7B size model further trained with DPO on top of Mistral, in Table~\ref{table: f1_score} show similar results; high degradation rate by DPO training.
 
\paragraph{Performance degradation is more influenced by the distraction level of the contexts rather than the length of distractor contexts}
\begin{figure}[ht!]
    \centering
    \begin{minipage}[b]{0.45\textwidth}
    \includegraphics[width=\textwidth]{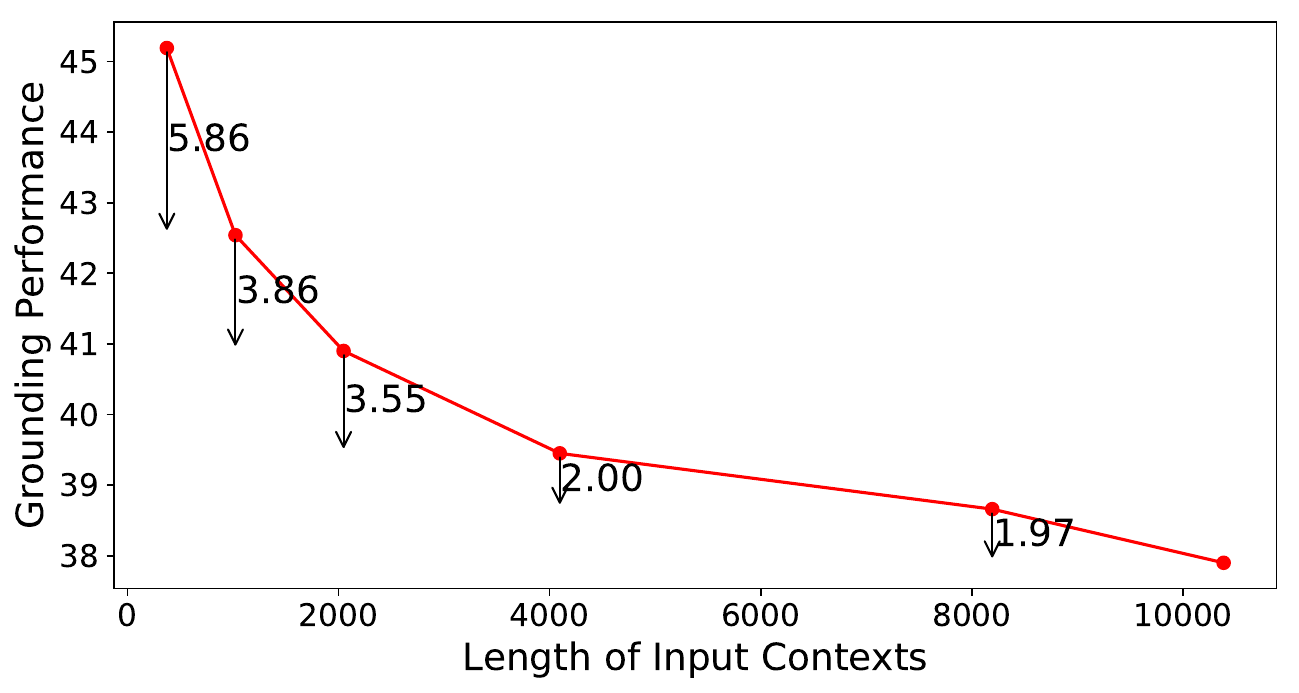}
    \caption{\fontsize{6.5}{10}\footnotesize Grounding performance of Vicuna-13B-16k as length of input contexts increases.} 
    \label{fig:dist_length}
    \end{minipage}
\end{figure}

Figure~\ref{fig:dist_length} illustrates that as the input context length increases, the grounding performance of Vicuna-13b-16k, capable of handling extensive inputs, varies significantly. Please note that the input contexts differ by the length of distractor contexts as the length of gold contexts is the same. 
Notably, grounding performance deteriorates more rapidly at the initial points (5.86 at the initial point and 1.97 at the end point of the plot).
This is because we add distractor contexts in the order of those in high rank by contriever~\citep{izacard2022unsupervised}, which indicates that contexts with high distraction levels are added at the initial points, causing stronger distractions. 
Such a result indicates that the performance decline is more influenced by the relevance and distraction level of the contexts, rather than the sheer number of distractors. 
The drop rate is mostly from the model's recall ability, highlighting its struggle to accurately identify all essential facts from the given contexts. 
This tendency shows a high correlation with a common challenge in retrieval models; performance decreases as they deal with larger data sets and encounter numerous query-relevant contexts within those sets~\citep{Zhong2023PoisoningRC}.

\paragraph{Impact of gold contexts position on grounding performance: optimal position at the end}
We could see that the position of gold contexts within multi-document settings significantly influences grounding performance, aligning with the findings from \citet{Liu2023LostIT}.
Experiment with Vicuna-13b-16k, input context length of 4096 over \textit{Original-dist} show the highest performance when gold contexts are positioned at the end and the lowest when positioned in the middle (end-43.37, beginning-39.32, random-39.45, middle-39.32).
The trend also holds for \textit{Conflict-dist}: end-43.53, beginning-41.28, random-39.10, middle-38.30. Such results emphasize the importance of where you put the gold contexts in a multi-document setting for high grounding performance. 

\paragraph{Instruction-tuned models show higher degradation with distractor contexts}
\begin{figure}[t!]
    \centering
    \begin{minipage}[b]{0.45\textwidth}
    \includegraphics[width=\textwidth]{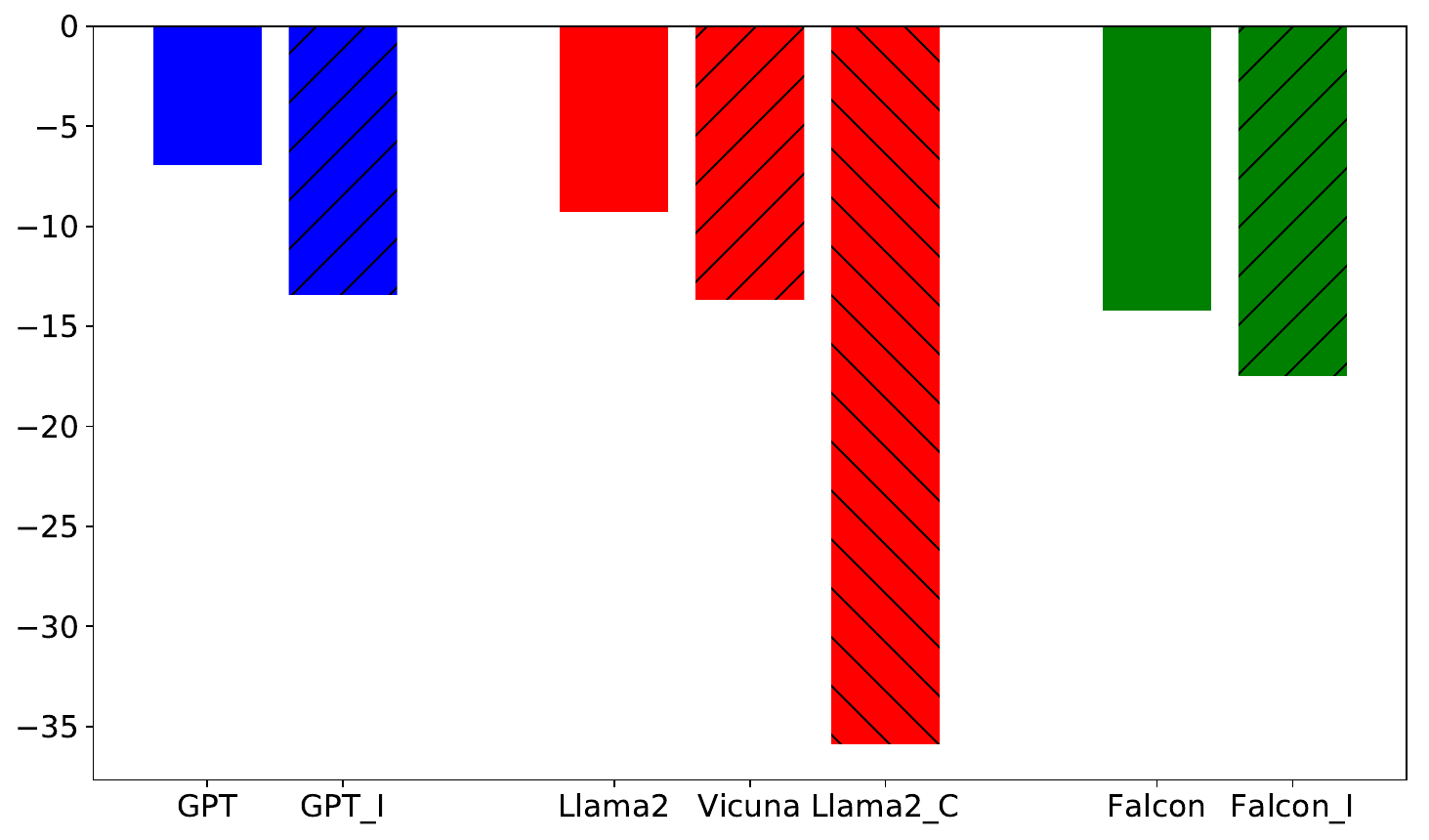}
    \caption{\fontsize{6.5}{10}\footnotesize Reduction rate in \textit{Original-Dist} performance from \textit{Original-Gold}. Models with the same base model are in the same color. Models that are instruction tuned (falcon\_I, GPT\_I, Vicuna) or underwent RLHF (Llama2\_C) show higher degradation when distractor contexts are added. Vicuna and Llama2 are 13B and Falcon is 40B model.}
    \label{fig:ori_dist}
    \end{minipage}
    \vspace{-1em}
\end{figure}

Figure~\ref{fig:ori_dist} demonstrates while models fine-tuned with instruction show higher absolute grounding performance, they show a notably greater decrease in performance when faced with distractor contexts. This trend is even more evident in models that underwent RLHF. We hypothesize that this decline in performance is likely a consequence of their tuning methods. During instruction tuning and RLHF, the models are trained to consider all input texts as relevant to their output generation. Consequently, they tend to incorporate distracting inputs when encountered. A closer examination of the metrics reveals a more pronounced drop in precision rather than recall. This suggests that in the presence of distractor contexts, these models are more inclined to use knowledge beyond the gold contexts, supporting our hypothesis. Thus, for instruction-tuned models, providing only the gold contexts without distractor contexts is crucial to maintain their high grounding performance.

\paragraph{Performance of answer accuracy}
Table~\ref{table: ans_acc} in Appendix~\ref{app: answer_acc} shows the answer accuracy of models across five settings. 
A key notable finding is that large-parameter models, like Falcon-40b, excel without contexts due to their inherent knowledge but see reduced gains with external contexts added as input. 
Also, without external contexts, high-popularity questions achieve a 32.6\% accuracy, outpacing low-popularity ones at 26.8\%. However, when with gold contexts: low-popularity questions slightly edge out at 83.4\% over the 83.2\% for high-popularity ones. We further analyze the generated response, we measure the fluency using G-EVAL~\citep{Liu2023GEvalNE} in Appendix~\ref{app: fluency}.

\section{Conclusion}
In this paper, we introduce a strict definition of ``grounding'' to external contexts when given as input.
To evaluate and analyze grounding performance under the definition, we propose a new dataset and grounding metric.
In our extensive evaluation of 25 LLMs across four dataset scenarios, we observed various insights. Rather than model size, various training techniques and base models tend to affect more on grounding performance. Models find it challenging to utilize all necessary knowledge when generating a response.
By presenting the performance of various models on different dataset settings, we provide valuable perspectives to the ongoing discourse on enhancing LLM grounding abilities and practical guidance for choosing suitable models for applications that require generating response by \textit{truly} grounding on a given context. 

\section{Limitations}

To construct a dataset with the specific requirements, all the contexts we utilize are sourced from Wikipedia, which is likely to be used as a source during pretraining LLMs. Therefore, to follow cases where private contexts (contexts that the model is likely to not have seen during training) we collect a modified version of the dataset, which also allows us to clearly differentiate between knowledge derived from the provided context and that inherent in the model's parameters. 
We leave collecting datasets with private contexts and evaluating the dataset as future work. As we modified the existing dataset, the contexts we provide may distract people.

While we have observed a high correlation with human judgments in our assessments, it's important to note that since our evaluation metric involves a model-based approach, the performance of the prediction model (\(M_{pred}\)) could be influenced by the performance of the evaluation model (\(M_{eval}\)). Therefore, the accuracy and reliability of \(M_{eval}\) are critical, as any limitations or biases within it could potentially affect the outcome of our performance evaluations for \(M_{pred}\). 
Additionally, while decomposing context into atomic facts also aligns well with human judgment, we note several failure cases attributable to model involvement, which further impacts grounding performance.

\section*{Acknowledgements}
This work was partly supported by Kakao Brain grant (2023, Aligning Language Model with Knowledge Module, 80\%) and Institute of Information \& communications Technology Planning \& Evaluation (IITP) grant funded by the Korea government (MSIT) (No.2022-0-00113, Developing a Sustainable Collaborative Multi-modal Lifelong Learning Framework, 20\%). 

\noindent We thank Seonghyeon Ye, Sewon Min, Yoonjoo Lee, Hanseok Oh, and Seungone Kim for helpful discussions and constructive feedback. We also thank Jonghyeon Kim, Daeyang Oh, Jungeun Lee, and Hyungyu Chae for annotating the data.

\bibliography{anthology,custom}
\bibliographystyle{acl_natbib}

\appendix

\section{Dataset Construction} \label{app: overall_data}
\begin{figure*}[t!]
    \centering
    \begin{minipage}[b]{0.99\textwidth}
    \includegraphics[width=\textwidth]{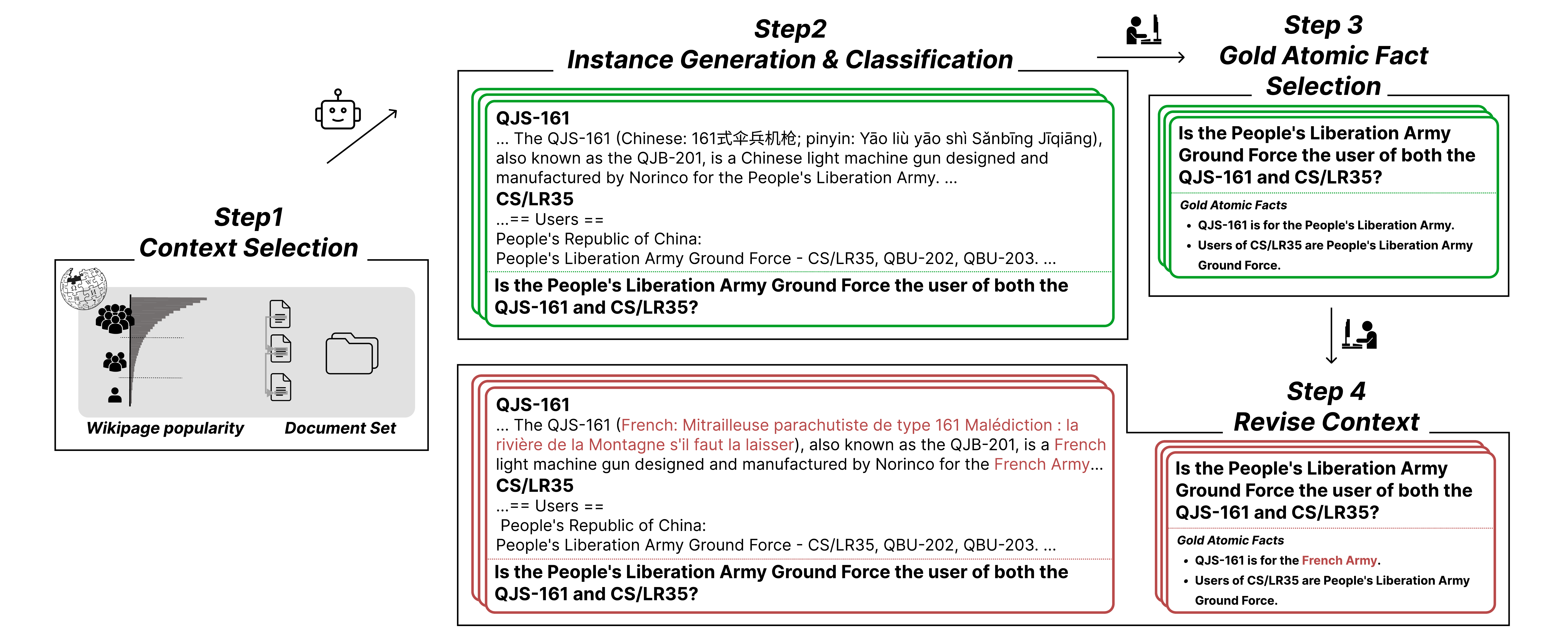}
    \caption{Data Construction Pipeline. Step 1-3 shows how we construct \textit{Original-Gold}, and Step 4 shows how we modified the dataset, thereby constructing \textit{Conflict-Gold}.}
    \label{fig:data_construction}
    \end{minipage}
\end{figure*}

\noindent As shown in Figure~\ref{fig:data_construction}, our dataset construction is mainly divided into four steps. Details of data construction including human annotators, inter-labeler agreement, data distribution of the factors, data examples, and more are in Appendix~\ref{app: overall_data}.

\subsection{[Step 1] Context Selection}

In the process of context selection, we focus on constructing a setup that reflects the popularity of the context topic and the required number of documents to answer the query. Wikipedia documents\footnote{Text in Wikipedia is co-licensed under the CC BY-SA and GFDL and is widely used in research.} were used for context, considering their comprehensive meta-information pertinent to these aspects. 
For Factor 1, we first start by quantifying the popularity of documents following \citet{Mallen2022WhenNT}. 
We calculate the sum of monthly pageviews\footnote{\url{https://dumps.wikimedia.org/other/pageview_complete/monthly/2023/}} for every six months from 2021 to 2023. From this, we derive a high and a low popularity list for the documents from the top and bottom 30\% range in consideration of Factor 1. 
Next, for Factor 2, each document within the popularity lists was grouped with additional documents retrieved through hyperlinks to make a document set. More specifically, an additional document was sampled from the intersection between the popularity list and hyperlinked document\footnote{It was observed that relevance between documents tends to diminish beyond three hyperlink hops; hence, we limited the document range from one to three hops.}. Such a process was done to construct a document set interconnected with each other, thus forming a comprehensive basis for generating queries requiring the integration of multiple sources as required for Factor 2.

\subsection{[Step 2] Detail of Instance Generation \& Classification}
\begin{figure*}[t!]
\centering
    \includegraphics[width=0.95\linewidth]{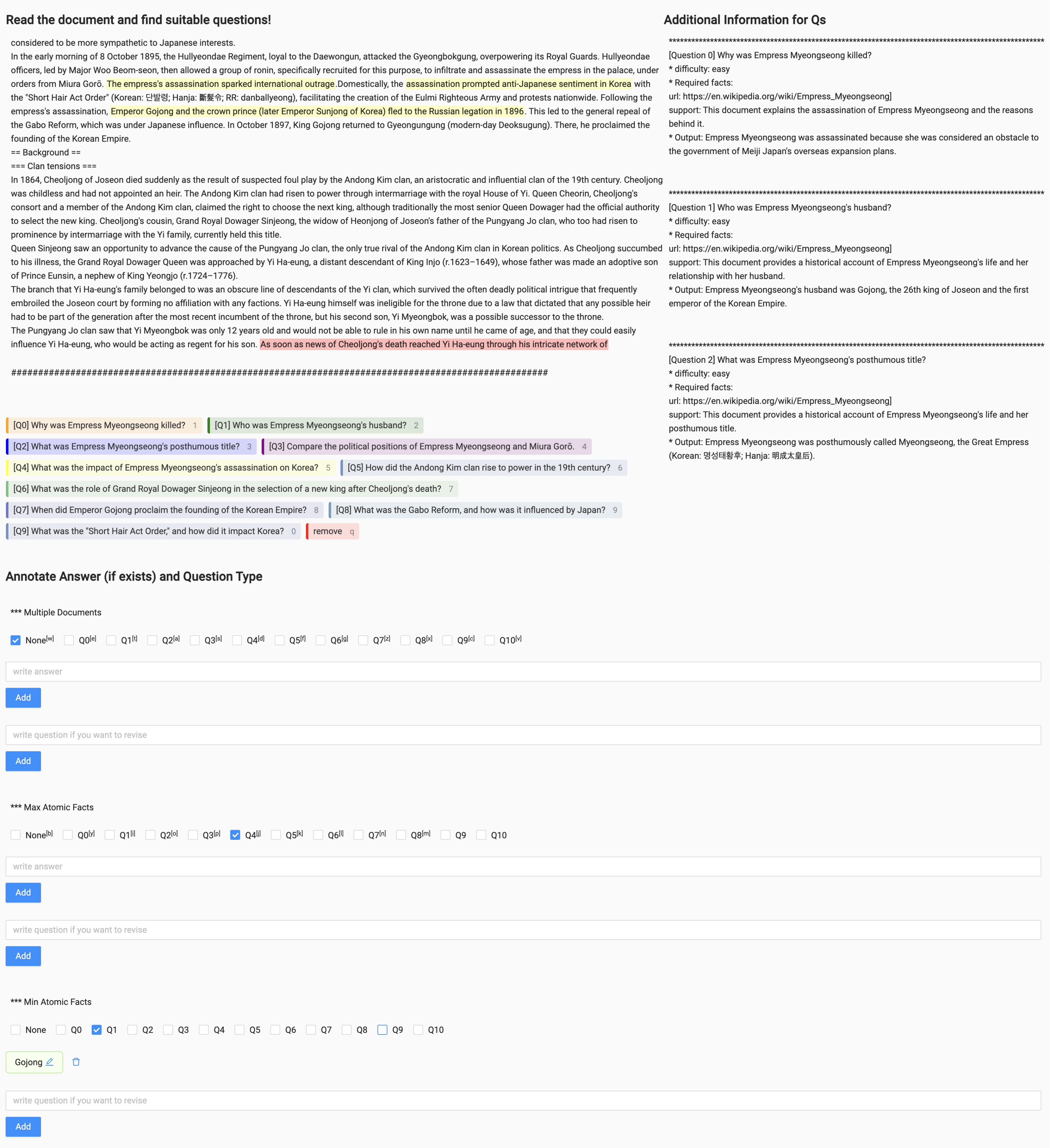}
    \caption{}
    \label{interface:instance}
\end{figure*}

Based on the document set from Step 1, we use ChatGPT to generate 10 candidate pairs of question and answer. Taking into account Factor 2 and Factor 3, we classify the generated queries on two criteria;  whether they require consideration of multiple contexts or single context (Factor 2) and whether they require a definite answer or free-form answer (Factor 3). 
During this classification process, pairs with low quality (e.g. meaningless conjunction of query from each document) or those requiring facts that don't exist in the given context are removed. 
Annotators label the minimal set out of the provided context to answer the question along with the span of context they used to generate an answer. 
During this process, annotators label the minimal set out of the provided context to answer the question. Annotators are asked to write all forms of answers
The interface used for instance filtering is in Figure~\ref{interface:instance}.
\label{appendix:wiki detail}

\subsection{[Step 3] Example of Atomic Facts}

For fine-grained evaluation, we decompose context sets into atomic facts. Atomic facts are short sentences conveying one piece of information. Following \citet{Min2023FActScoreFA}, we use InstructGPT to decompose. Example results of atomic facts decomposed when given a sentence is in Table~\ref{table: atomic_facts_example}.

\label{appendix: atomic facts}
\begin{table*}[t!]
\centering
\fontsize{7.5}{10}\selectfont
\caption{\fontsize{7.5}{10}\footnotesize Examples of Atomic Facts for each sentence.} 
\begin{tabular}{ m{6cm} m{9cm}} 
    \toprule
    \textbf{Sentence} & \textbf{Atomic Facts} \\
    \midrule
        \multirow{5}{=}{\\[1em] The Indian Premier League (IPL) (also known as the TATA IPL for sponsorship reasons) is a men's Twenty20 (T20) cricket league that is annually held in India and contested by ten city-based franchise teams.}
        & \textbf{Fact 1:} The Indian Premier League is a men's Twenty20 cricket league.\\
        \cmidrule(lr){2-2}
        & \textbf{Fact 2:} The Indian Premier League is annually held in India.\\
        \cmidrule(lr){2-2}
        & \textbf{Fact 3:} The Indian Premier League is contested by ten city-based franchise teams.\\
        \cmidrule(lr){2-2}
        & \textbf{Fact 4:} The Indian Premier League is also known as the TATA IPL.\\
        \cmidrule(lr){2-2}
        & \textbf{Fact 5:} The Indian Premier League is known as the TATA IPL for sponsorship reasons.\\
    \midrule
        \multirow{3}{=}{\\[0.3em] The league's format was similar to that of the English Premier League  and the National Basketball Association in the United States.}
        & \textbf{Fact 1:} The league had a format.\\
        \cmidrule(lr){2-2}
        & \textbf{Fact 2:} The league's format was similar to the English Premier League.\\
        \cmidrule(lr){2-2}
        & \textbf{Fact 3:} The league's format was similar to the National Basketball Association in the United States.\\
    \midrule
        \multirow{4}{=}{\\[1em] The Indian Cricket League (ICL) was founded in 2007 with funding provided by Zee Entertainment Enterprises.}
        & \textbf{Fact 1:} The Indian Cricket League (ICL) was founded.\\
        \cmidrule(lr){2-2}
        & \textbf{Fact 2:} The Indian Cricket League (ICL) was founded in 2007.\\
        \cmidrule(lr){2-2}
        & \textbf{Fact 3:} Funding was provided for the founding of the Indian Cricket League (ICL).\\
        \cmidrule(lr){2-2}
        & \textbf{Fact 4:} Zee Entertainment Enterprises provided funding for the founding of the Indian Cricket League (ICL).\\
    \midrule
        \multirow{4}{=}{\\[1em] The first season was due to start in April 2008 in a 'high-profile ceremony' in New Delhi.}
        & \textbf{Fact 1:} The first season was due to start.\\
        \cmidrule(lr){2-2}
        & \textbf{Fact 2:} The first season was due to start in April 2008.\\
        \cmidrule(lr){2-2}
        & \textbf{Fact 2:} The first season was due to start in a high-profile ceremony.\\
        \cmidrule(lr){2-2}
        & \textbf{Fact 2:} The high-profile ceremony was in New Delhi.\\
    \bottomrule
\end{tabular}
\label{table: atomic_facts_example}
\end{table*}

\subsection{[Step 3] Gold Atomic Annotation Interface}
\begin{figure*}[t!]
\centering
    \includegraphics[width=0.95\linewidth]{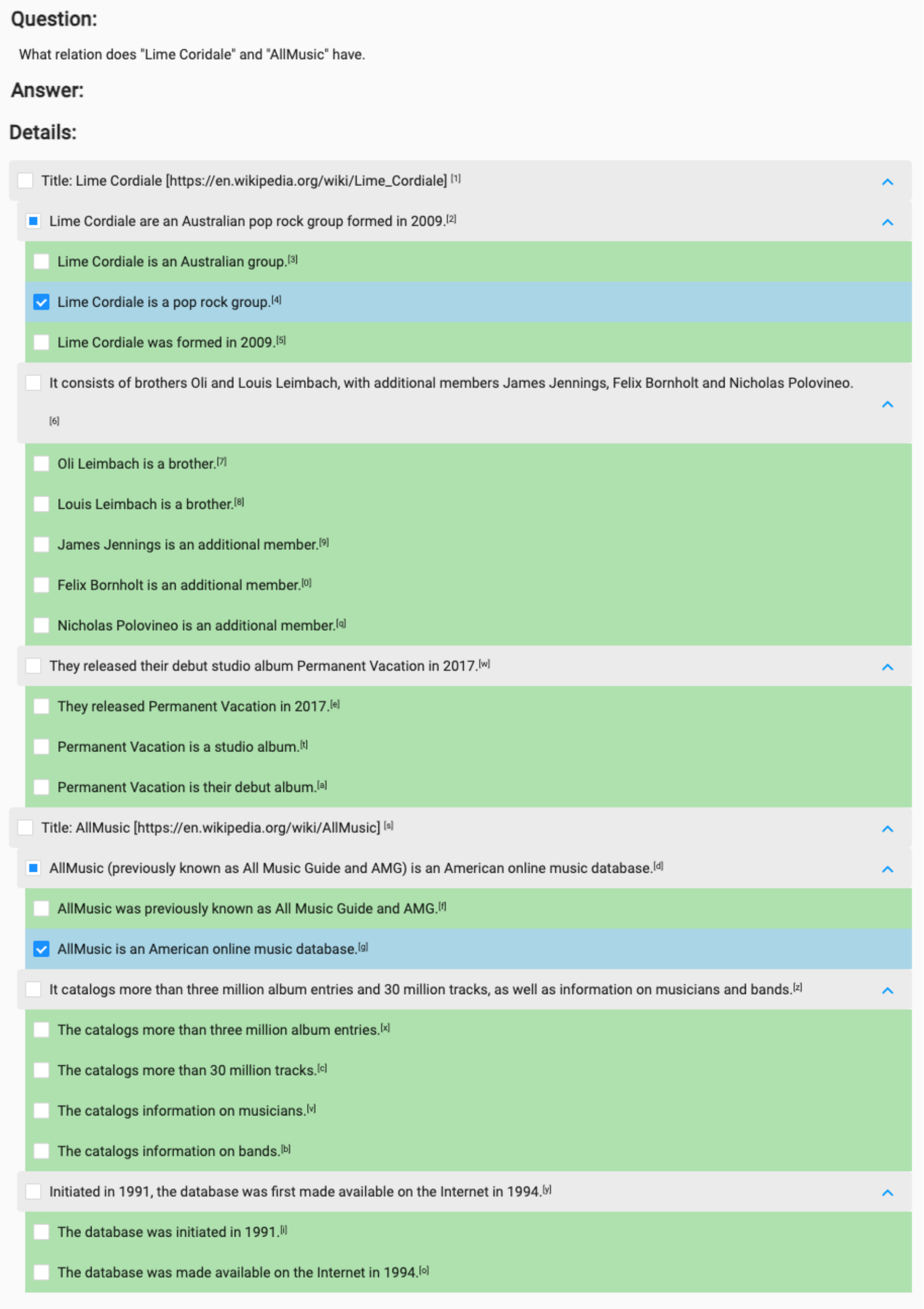}
    \caption{User interface used for gold atomic annotation}
    \label{interface:godl_atomic_annotation}
\vspace{-3mm}
\end{figure*}

From the atomic facts, we further annotate the gold ones, which we call gold atomic facts. Figure~\ref{interface:godl_atomic_annotation} is the interface used to annotate gold atomic facts.
We get a high correlation between annotators; 0.82 when calculated with Cohen's Kappa.

\label{appendix: gold_atomic_interface}

\subsection{[Step 4] Modify Context Interface}

\begin{figure*}[t!]
\centering
    \includegraphics[width=0.95\linewidth]{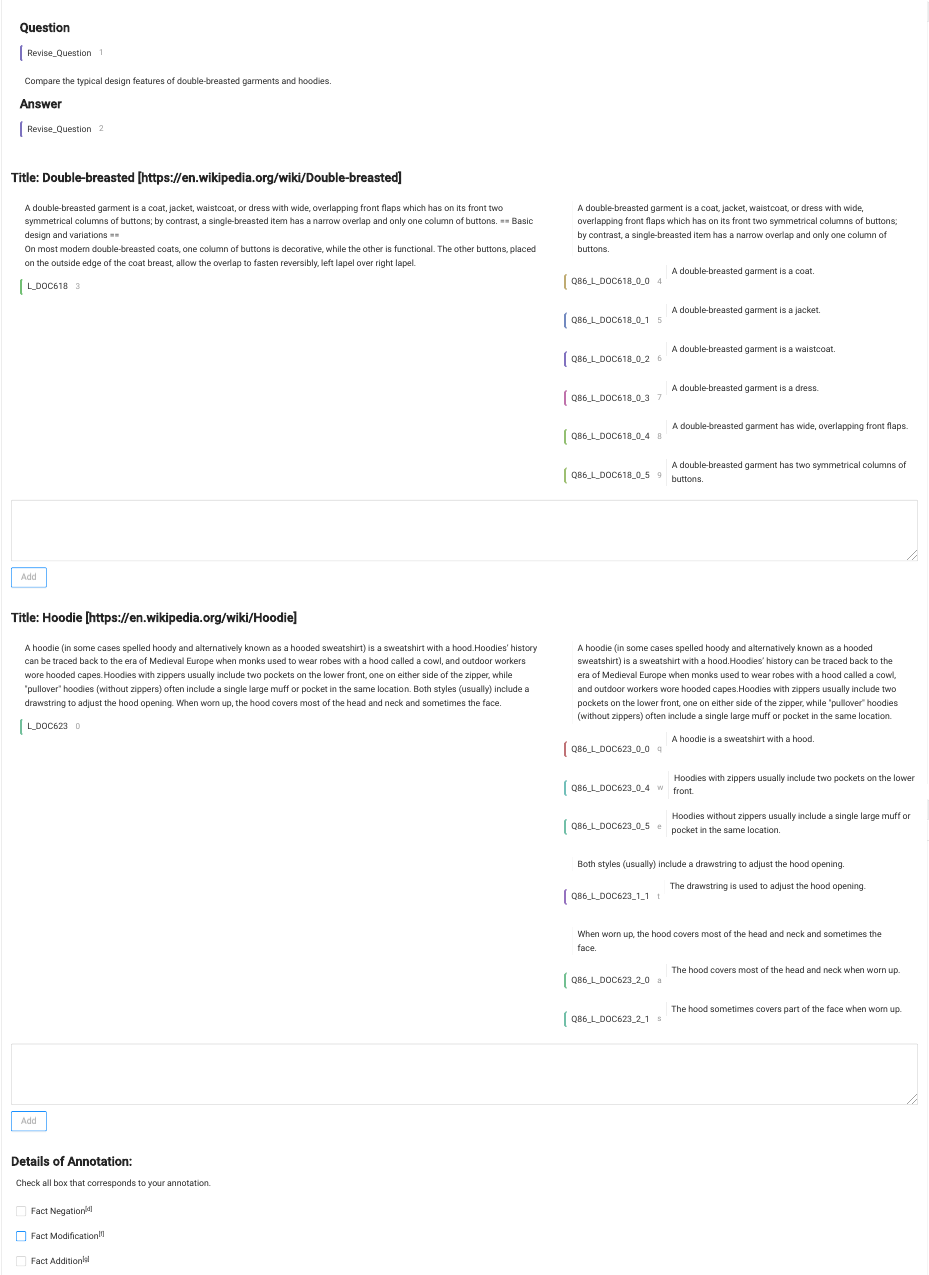}
    \caption{An illustration of the interface to modify context. The question, answer, input context, and corresponding gold atomics are given to the annotators and annotators should modify well-known information by revising gold atomic facts and input contexts. Annotators are also asked to check which type of modification they did.}
    \label{interface:revised}
\end{figure*}

Human annotators are told to revise the instance in a way that they would be wrong if they had answered the question based on background knowledge, not based on the input context. Revision to any part of the instance was applied across the whole instance. For instance, if a fact negation was done on an atomic fact, any related parts of the question, context, and answer were also negated. 
The purpose of such instructions was to generate an instance with gold atomic facts that are unlikely to be found in the pretrained dataset, thereby distinguishing information from its parametric space. 
Figure~\ref{interface:revised} is the interface used to construct a modified version of the dataset.

\label{appendix: revise_context_interface}

\subsection{Human Annotators}  \label{app: human_labelers}
We recruit 4 Korean college students proficient in English and pay \$15 USD per hour for step 4. 
The annotation was done in a two-phase process. Initially, the annotators dedicated 1.5 hours to the task, after which they received guidance on any errors made before completing the remaining annotations. For the rest of the steps, the authors took part in the annotation process.

\subsection{Data Distribution} \label{app: data_distribution}
After following the dataset construction step, we have 480 datasets (question, answer, context, gold atomic facts) along with 480 modified context pairs. 
In terms of distribution characteristics, we aimed to balance the various factors. Specifically, for Factor 1 and Factor 3, we achieve an approximate 50\% distribution for both high (53.3\%) and low (46.7\%) popularity levels and for definite (54.1\%) and free-form (45.9\%) answer types. However, concerning Factor 2, which revolves around the source multiplicity of our queries, it was challenging to generate high-quality queries from multiple sources in Step 2, thereby only 16.7\% of the queries derived from multiple sources, with a predominant 83.3\% stemming from a single source.

\subsection{Dataset Examples} \label{appendix: data example}
Table~\ref{example: instances} shows examples of instances within the new dataset we propose. 

\begin{table*}
\centering
\fontsize{7.5}{10}\selectfont
\begin{tblr}{
  width = \linewidth,
  colspec = {Q[152]Q[480]Q[225]Q[57]},
  row{1} = {c},
  cell{2}{4} = {c},
  cell{3}{1} = {r=2}{},
  cell{3}{4} = {r=2}{},
  hline{1,6} = {-}{0.08em},
  hline{2-3,5} = {-}{0.05em},
  hline{4} = {2-3}{},
}
\textbf{Question} & \textbf{Context} & \textbf{Gold Atomic} & \textbf{Answer}\\
Provide the claimed number of Viet Cong killed during Operation Sunset Beach. & {\textbf{Operation Sunset Beach ::} On 20 September the 1st Battalion, 5th Infantry Regiment (Mechanized) conducted a sweep of the Boi Loi Woods, meeting sporadic resistance and destroying bunkers and supplies.\\== Aftermath ==\\Operation Sunset Beach officially concluded on 11 October, with US reports claiming that \uline{Viet Cong losses were 80 killed (body count) and a further 135 estimated killed, U.S. losses were 29 killed. }\\== References ==\\~This article incorporates public domain material from websites or documents of the United States Army Center of Military History. } & {\labelitemi\hspace{\dimexpr\labelsep+0.5\tabcolsep}US reports claim Viet Cong losses were 80 killed (body count).~\\\labelitemi\hspace{\dimexpr\labelsep+0.5\tabcolsep}US reports estimate Viet Cong losses were 135 killed.} & 215\\
What manufacturer provided the v8 engine that went into the Holden designed model which ceased production on 20 October 2017. & \textbf{\textbf{Holden ::}}~On 29 November 2016, engine production at the Fishermans Bend plant was shut down. \uline{On 20 October 2017, production of the last Holden designed Commodore ceased and the vehicle assembly plant at Elizabeth was shut down. }Holden produced nearly 7.7 million vehicles. & \labelitemi\hspace{\dimexpr\labelsep+0.5\tabcolsep}On 20 October 2017, production of the last Holden designed Commodore ceased. & Chevrolet\\
 & \textbf{\textbf{Holden Commodore (VX) ::~}}The optional Supercharged Ecotec V6 extended its service to the Executive and Acclaim variants, with the 171-kilowatt (229 hp) output figure remaining unchanged from the VT. As well as the supercharged six-cylinder, an even more powerful \uline{5.7-litre Chevrolet-sourced Gen III V8 engine was offered}. The powerplant received power increases from 220 to 225 kilowatts (295 to 302 hp). A modified front suspension setup received lower control arm pivot points. The Series II update featured the addition of a new rear cross member, revised rear control arm assemblies with new style bushing and toe-control links to the semi-trailing arm rear suspension to better maintain the toe settings during suspension movements, resulting in more predictable car handling, noticeably over uneven surfaces, and improved tyre wear. & {\labelitemi\hspace{\dimexpr\labelsep+0.5\tabcolsep}The 5.7-litre engine was Chevrolet-sourced. \\\labelitemi\hspace{\dimexpr\labelsep+0.5\tabcolsep}The 5.7-litre engine was a Gen III V8.} & \\
Explain what a "dump" refers to in volleyball. & {\textbf{Volleyball jargon :: }Arms can be in a platform position or in a overhead position like a set. The player digs the ball when it is coming at a downward trajectory\\Double contact or Double touch: A fault in which a player contacts the ball with two body parts consecutively\\D.S. : The abbreviation for "defensive specialist", a position player similar to the libero who is skilled at back row defense\\\uline{Dump: A surprise attack usually executed by a front row setter to catch the defense off guard; many times executed with the left hand, sometimes with the right, aimed at the donut or area 4 on the court.}\\Five-One: Six-player offensive system where a single designated setter sets regardless of court position.     } & {\labelitemi\hspace{\dimexpr\labelsep+0.5\tabcolsep}A dump is a surprise attack.~\\\labelitemi\hspace{\dimexpr\labelsep+0.5\tabcolsep}A dump is usually executed by a front row setter.~\\\labelitemi\hspace{\dimexpr\labelsep+0.5\tabcolsep}A dump is executed to catch the defense off guard.~\\\labelitemi\hspace{\dimexpr\labelsep+0.5\tabcolsep}A dump is sometimes executed with the left hand.~\\\labelitemi\hspace{\dimexpr\labelsep+0.5\tabcolsep}A dump is sometimes executed with the right hand.~\\\labelitemi\hspace{\dimexpr\labelsep+0.5\tabcolsep}A dump is aimed at the donut or area 4 on the court.} & 
\end{tblr}

\caption{Example of Instances}
\label{example: instances}
\end{table*}

\begin{table*}
\centering
\fontsize{7.5}{10}\selectfont
\begin{tblr}{
  width = \linewidth,
  colspec = {Q[152]Q[480]Q[225]Q[57]},
  row{1} = {c},
  cell{2}{4} = {c},
  cell{3}{1} = {r=2}{},
  cell{3}{4} = {r=2}{},
  hline{1,6} = {-}{0.08em},
  hline{2-3,5} = {-}{0.05em},
  hline{4} = {2-3}{},
}
\textbf{Question} & \textbf{Context} & \textbf{Gold Atomic} & \textbf{Answer}\\
Provide the claimed number of Viet Cong killed during Operation Sunset Beach. & {\textbf{Operation Sunset Beach ::} On 20 September the 1st Battalion, 5th Infantry Regiment (Mechanized) conducted a sweep of the Boi Loi Woods, meeting sporadic resistance and destroying bunkers and supplies.\\== Aftermath ==\\Operation Sunset Beach officially concluded on 11 October, with US reports claiming that \uline{Viet Cong losses were \textit{180} killed (body count) and a further \textit{235} estimated killed, U.S. losses were 29 killed. }\\== References ==\\~This article incorporates public domain material from websites or documents of the United States Army Center of Military History. } & {\labelitemi\hspace{\dimexpr\labelsep+0.5\tabcolsep}US reports claim Viet Cong losses were \textit{180} killed (body count).~\\\labelitemi\hspace{\dimexpr\labelsep+0.5\tabcolsep}US reports estimate Viet Cong losses were \textit{235} killed.} &\textit{415}\\
What manufacturer provided the v8 engine that went into the Holden designed model which ceased production on 20 October 2017. & \textbf{\textbf{Holden ::}}~On 29 November 2016, engine production at the Fishermans Bend plant was shut down. \uline{On 20 October 2017, production of the last Holden designed Commodore ceased and the vehicle assembly plant at Elizabeth was shut down. }Holden produced nearly 7.7 million vehicles. & \labelitemi\hspace{\dimexpr\labelsep+0.5\tabcolsep}On 20 October 2017, production of the last Holden designed Commodore ceased. & \textit{Audi}\\
 & \textbf{\textbf{Holden Commodore (VX) ::~}}The optional Supercharged Ecotec V6 extended its service to the Executive and Acclaim variants, with the 171-kilowatt (229 hp) output figure remaining unchanged from the VT. As well as the supercharged six-cylinder, an even more powerful \uline{5.7-litre \textit{Audi-sourced} Gen III V8 engine was offered}. The powerplant received power increases from 220 to 225 kilowatts (295 to 302 hp). A modified front suspension setup received lower control arm pivot points. The Series II update featured the addition of a new rear cross member, revised rear control arm assemblies with new style bushing and toe-control links to the semi-trailing arm rear suspension to better maintain the toe settings during suspension movements, resulting in more predictable car handling, noticeably over uneven surfaces, and improved tyre wear. & {\labelitemi\hspace{\dimexpr\labelsep+0.5\tabcolsep}The 5.7-litre engine was \textit{Audi-sourced}. \\\labelitemi\hspace{\dimexpr\labelsep+0.5\tabcolsep}The 5.7-litre engine was a Gen III V8.} & \\
Explain what a "dump" refers to in volleyball. & {\textbf{Volleyball jargon :: }Arms can be in a platform position or in a overhead position like a set. The player digs the ball when it is coming at a downward trajectory\\Double contact or Double touch: A fault in which a player contacts the ball with two body parts consecutively\\D.S. : The abbreviation for "defensive specialist", a position player similar to the libero who is skilled at back row defense\\\uline{Dump: A \textit{final blow} usually executed by a front row setter to catch the defense off guard; many times executed with the left hand, sometimes with the right, aimed at the donut or area 4 on the court.}\\Five-One: Six-player offensive system where a single designated setter sets regardless of court position.     } & {\labelitemi\hspace{\dimexpr\labelsep+0.5\tabcolsep}A dump is a \textit{final blow}.~\\\labelitemi\hspace{\dimexpr\labelsep+0.5\tabcolsep}A dump is usually executed by a front row setter.~\\\labelitemi\hspace{\dimexpr\labelsep+0.5\tabcolsep}A dump is executed to catch the defense off guard.~\\\labelitemi\hspace{\dimexpr\labelsep+0.5\tabcolsep}A dump is sometimes executed with the left hand.~\\\labelitemi\hspace{\dimexpr\labelsep+0.5\tabcolsep}A dump is sometimes executed with the right hand.~\\\labelitemi\hspace{\dimexpr\labelsep+0.5\tabcolsep}A dump is aimed at the donut or area 4 on the court.} & 
\end{tblr}
\label{example: revised instances}
\caption{Example of Modified Instances}
\end{table*}

\subsection{Adding Distractor Context}
\label{appendix: distractor_context}
We employ contriever~\cite{izacard2022unsupervised}, a dense retriever pretrained through contrastive learning, to retrieve the top 40 contexts with high similarity to each question from the corpus used in our benchmark. 
Please note that for each question, we exclude contexts from Wikipedia documents that contain gold atomic facts due to the concern about potential changes or additions to these gold atomic facts. Examples of distractor contexts are in Table~\ref{table: distractor_examples}.

\begin{table*}[t!]
\centering
\fontsize{7.5}{9}\selectfont
\caption{\fontsize{7.5}{10}\footnotesize Examples of Distractor Contexts.} 
\begin{tabular}{ m{4cm} m{6cm} m{6cm}} 
    \toprule
    \textbf{Question} & \textbf{Gold Context} & \textbf{Distractor Context} \\
    \midrule
        \multirow{15}{=}{\\[1em] What is a common factor of Sepsis and Hypotension?}
        & \textbf{Title:} Sepsis\newline \textbf{Context:} Sepsis (septicaemia in British English), or blood poisoning, is a life-threatening condition that arises when the body's response to infection causes injury to its own tissues and organs.This initial stage of sepsis is followed by suppression of the immune system. Common signs and symptoms include fever, increased heart rate, increased breathing rate, and confusion. There may also be symptoms related to a specific infection, such as a cough with pneumonia, or painful urination with a kidney infection. \newline\newline \textbf{Title:} Hypotension\newline \textbf{Context:} Hypotension is low blood pressure. Blood pressure is the force of blood pushing against the walls of the arteries as the heart pumps out blood. Blood pressure is indicated by two numbers, the systolic blood pressure (the top number) and the diastolic blood pressure (the bottom number), which are the maximum and minimum blood pressures, respectively. & \textbf{\#Top1}\newline \textbf{Title:} Gunshot wound\newline \textbf{Context:} Long-term complications can include bowel obstruction, failure to thrive, neurogenic bladder and paralysis, recurrent cardiorespiratory distress and pneumothorax, hypoxic brain injury leading to early dementia, amputations, chronic pain and pain with light touch (hyperalgesia), deep venous thrombosis with pulmonary embolus, limb swelling and debility, lead poisoning, and post-traumatic stress disorder (PTSD). Factors that determine rates of gun violence vary by country. These factors may include the illegal drug trade, easy access to firearms, substance misuse including alcohol, mental health problems, firearm laws, social attitudes, economic differences and occupations such as being a police officer. Where guns are more common, altercations more often end in death. Before management begins it should be verified the area is safe. \\
        \cmidrule(lr){3-3}
        & & \textbf{\#Top2}\newline \textbf{Title:} Medical glove\newline \textbf{Context:} Medical gloves are recommended to be worn for two main reasons: To reduce the risk of contamination of health-care workers hands with blood and other body fluids. To reduce the risk of germ dissemination to the environment and of transmission from the health-care worker to the patient and vice versa, as well as from one patient to another.\newline== History ==\newline Caroline Hampton became the chief nurse of the operating room when Johns Hopkins Hospital opened in 1889. \\ 
        \cmidrule(lr){3-3}
        & & $$\vdots$$ \\
    \midrule
        \multirow{12}{=}{\\[1em] What was the initial name of .223 Remington?}
        & \textbf{Title:} .223 Remington\newline \textbf{Context:} This cartridge is loaded with DuPont IMR4475 powder.During parallel testing of the T44E4 (future M14) and the ArmaLite AR-15 in 1958, the T44E4 experienced 16 failures per 1,000 rounds fired compared to 6.1 for the ArmaLite AR-15. Because of several different .222 caliber cartridges that were being developed for the SCHV project, the .222 Special was renamed .223 Remington. In May 1959, a report was produced stating that five- to seven-man squads armed with ArmaLite AR-15 rifles have a higher hit probability than 11-man squads armed with the M-14 rifle. & \textbf{\#Top1}\newline \textbf{Title:} .35 Remington\newline \textbf{Context:} The .35 Remington (9.1 x 49 mm) is the only remaining cartridge from Remington's lineup of medium-power rimless cartridges still in commercial production. Introduced in 1906, it was originally chambered for the Remington Model 8 semi-automatic rifle in 1908.It is also known as 9 x 49 mm Browning and 9 mm Don Gonzalo. == History ==\newline Over the years, the .35 Remington has been chambered in a variety of rifles by most firearms manufacturers, and continues in popularity today in the Marlin Model 336 lever-action and Henry Side Gate Lever Action. \\
        \cmidrule(lr){3-3}
        & & \textbf{\#Top2}\newline \textbf{Title:} Squad automatic weapon\newline \textbf{Context:} During its long service in the US military, it was pivotal in the evolution of U.S. fireteam tactics and doctrine that continues to the present day. Modern squad automatic weapons (such as the RPK and L86) are modified assault rifles or battle rifles (e.g. FN FAL 50.41 and M14A1) that may have increased ammunition capacity and heavier barrels to withstand continued fire and will almost always have a bipod. In the case of some assault rifles, such as the H\&K G36 or Steyr AUG, the SAW is simply the standard rifle with a few parts replaced. \\
        \cmidrule(lr){3-3}
        & & $$\vdots$$ \\
    \bottomrule
\end{tabular}
\label{table: distractor_examples}
\end{table*}

\subsection{Difference from existing datasets}
Our dataset differs from previous knowledge retrieval datasets in three key aspects. 
First is the existence of gold atomic facts annotation. Gold atomic facts are necessary to calculate recall performance; as previous works focused on calculating only precision, there is no dataset with gold atomic facts annotation.
The second is conflict QA pair inclusion. Our dataset contains conflict QA pairs to differentiate between knowledge derived from external sources and memorized knowledge; to see whether the model generates a response by truly grounding on external context rather than generating a memorized one.
Last is the consideration of multiple axes. We take into account various axes ($\mathcal{F}_1$, $\mathcal{F}_2$, $\mathcal{F}_3$in Figure 2) widely recognized to impact knowledge augmented LM performance together.
Table~\ref{table: diff} shows the clear distinctions between our dataset and others for a comprehensive understanding.

\begin{table*}[h]
\centering
\fontsize{6.5}{10}\selectfont
    \begin{tabular}{c|ccccccc}
    \toprule
    Datasets & 	Knowledge Conflict & NQ & HotpotQA & StrategyQA & popQA & Factscore& Ours \\
    \midrule
    Annotation of gold atomic facts & x & x& x & x&x&x&o \\
    Existence of conflict QA pair& x & x& x & x&x&x&o \\
    Existence of gold context pair to answer&	o	&o	&o&	x	&o	&x	&o\\
    Consideration of popularity ($\mathcal{F}_1$) & o&	x&	x&	x&	o&	o	&o \\
    Range of number of contexts ($\mathcal{F}_2$) & 1	&1	&2	&-	&-	&1&	1-3\\
    Response format ($\mathcal{F}_3$)&	definite&	definite&	definite&	definite&	definite&	free-form&	both\\

    \bottomrule
    \end{tabular}
\caption
     {\fontsize{6.5}{10}\footnotesize Comparison with widely used datasets: knowledge conflict~\citep{faille-etal-2021-entity-based}, NQ~\citep{kwiatkowski-etal-2019-natural}, HotpotQA~\citep{yang2018hotpotqa}, StrategyQA~\citep{geva2021did}, popQA~\citep{Mallen2022WhenNT}, and Factscore~\citep{Min2023FActScoreFA}.} 
     \vspace{-1em}
\label{table: diff}
\end{table*}

\section{Evaluate Human Correlation for $M_{eval}$}
\label{appendix: human_eval}

As the same knowledge could be represented in various ways, we utilize a prediction model $M_{eval}$, which predicts whether knowledge of each atomic fact is in a generated response or input context. We evaluate five different $M_{eval}$ and choose the one with the highest correlation with humans. In section~\ref{app: human_eval_interface}, we show the interface we used by human evaluators. In section~\ref{app: human_eval_details}, we share the details on the models we used and how we used them.

We assess the presence of the knowledge by evaluation model ($M_{eval}$) as the same information can be expressed in various ways; $M_{eval}$ evaluates whether an atomic fact is in the given information. 
Since grounding performance can vary depending on the performance of $M_{eval}$, we conduct evaluations using five different models\footnote{Details of the models are in Appendix~\ref{appendix: human_eval}.} and utilize the one with the highest correlation with human evaluation as $M_{eval}$. 
As shown in Figure~\ref{table: human_correlation}, the cross-encoder model trained on MSMARCO dataset\footnote{cross-encoder/ms-marco-MiniLM-L-12-v2 from Sentence Transformers~\citep{reimers-2019-sentence-bert}} shows the highest correlation with humans. 
This model not only surpasses GPT4 in terms of correlation but also demonstrates a correlation metric analogous to human-to-human correlation (88.6). Given these findings, we have chosen to employ the cross-encoder model as our evaluation model ($M_{eval}$).

\subsection{Human Evaluation Interface} \label{app: human_eval_interface}
\begin{figure*}[t!]
\centering
    \includegraphics[width=0.89\linewidth]{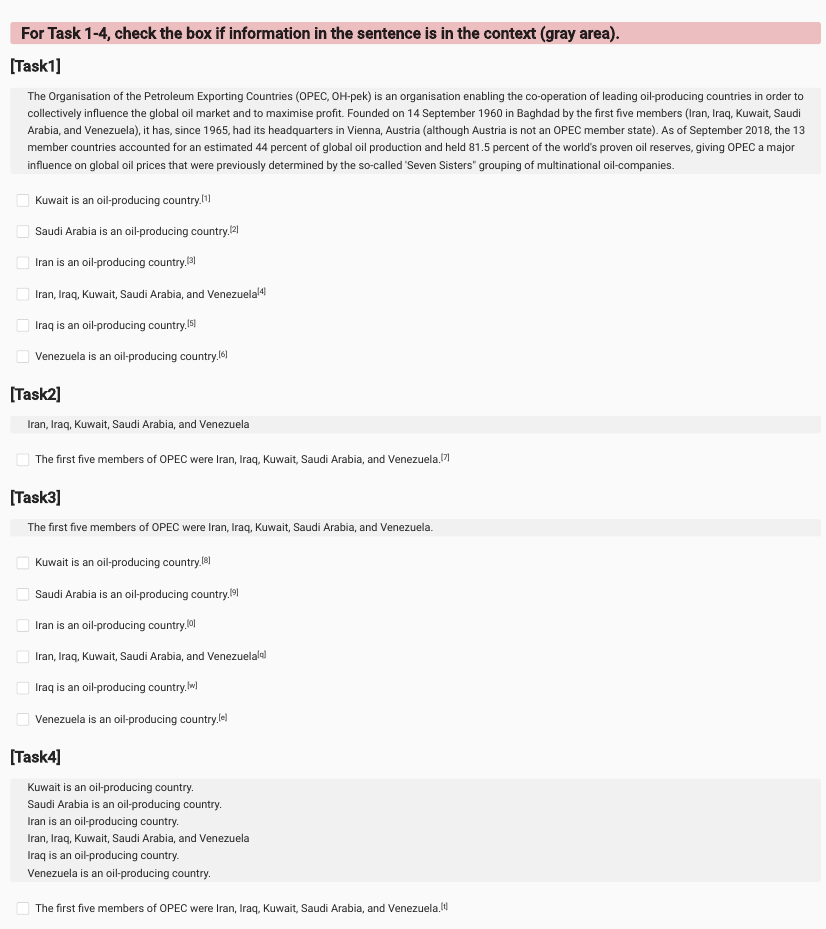}
    \caption{An illustration of the human evaluation to calculate the correlation with $M_{eval}$. Task 1 and Task 2 are to evaluate correlation with $\text{GR}_\text{loose}$, which is to check whether the given atomic fact is in the paragraph, and Task 3 and Task4 are to evaluate correlation with $\text{GR}_\text{strict}$, which is to compare between the atomic facts.}
    \label{interface:human_eval}
\end{figure*}

Figure~\ref{interface:human_eval} shows the interface used by human evaluators.
Humans are asked to evaluate whether the given atomic fact is in the context, the same operation as $M_{eval}$.
The inter-annotator-agreement (IAA) score is 88.6.

\subsection{Details of $M_{eval}$} \label{app: human_eval_details}

\begin{figure}[htb!]
    \centering
    \begin{minipage}[b]{0.45\textwidth}
    \includegraphics[width=\textwidth]{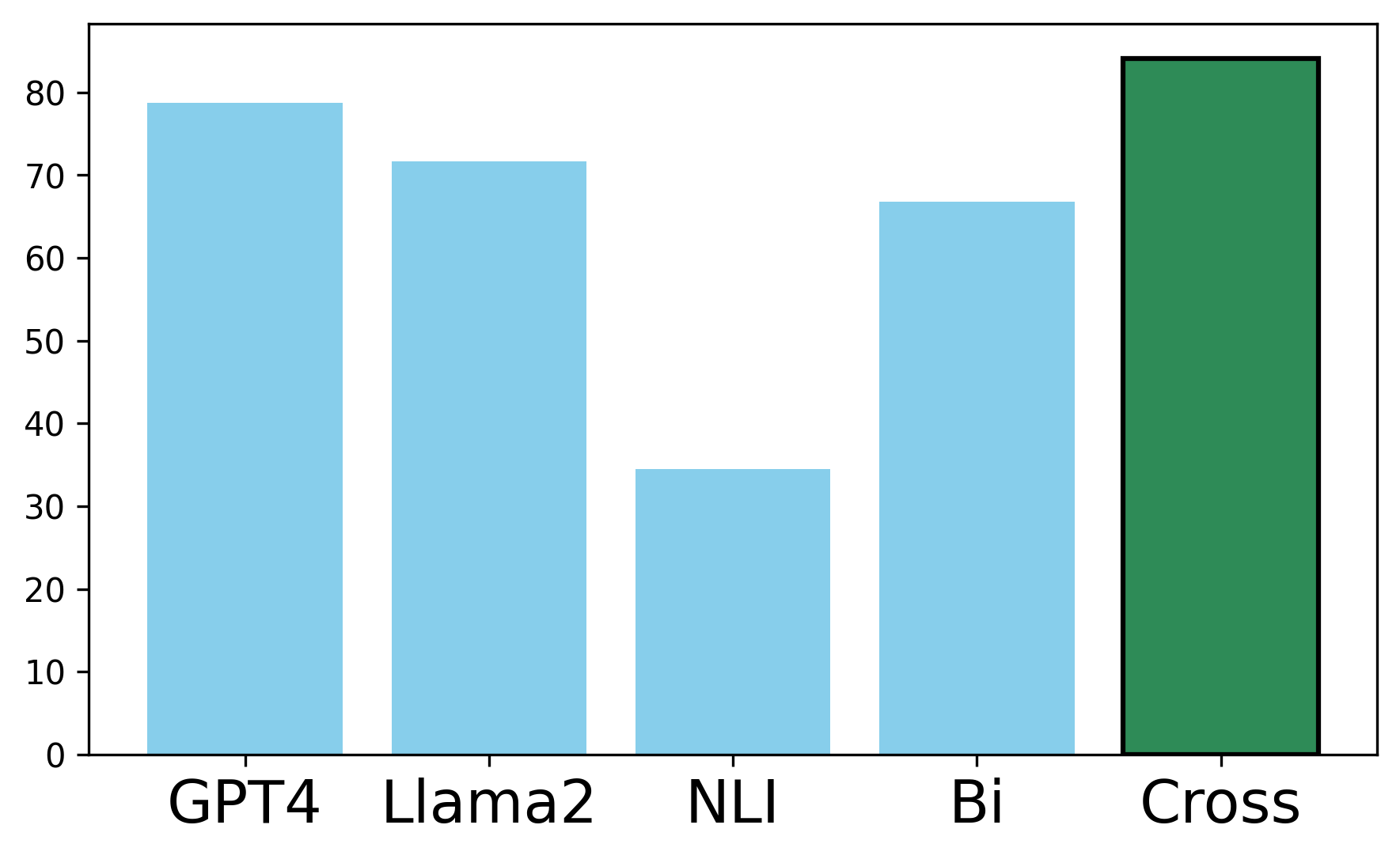}
    \caption{Correlation between Human and five models ($M_{eval}$) on predicting whether the knowledge of atomic facts are in a paragraph}
    \label{table: human_correlation}
    \end{minipage}
\end{figure}

\paragraph{GPT4, Llama-2-Chat-70b}
For GPT4 and Llama-70b-chat, same instruction is given following \citet{Min2023FActScoreFA} to evaluate:
\begin{boxC}
* context: \{\textit{paragraph}\}\newline\newline
* statement: \{\textit{atomic fact}\}\newline\newline
Generate 'True' if all information in given statement is in given context. Else generate 'False'
\end{boxC}
\paragraph{NLI}
For the NLI model, we use TRUE, a T5-XXL model trained on multiple NLI datasets. It has shown high performance in predicting whether the statement entails the other statement. We used the checkpoint released from \href{https://huggingface.co/google/t5_xxl_true_nli_mixture}{huggingface}. 

\paragraph{Bi, Cross}

To discern the presence of specific atomic facts within the provided contexts or generated responses, we adopted a text similarity-based methodology. By computing similarity scores between atomic facts and the context or responses, we can determine the inclusion or exclusion of certain knowledge segments.
In the pursuit of deriving robust similarity metrics, we opted for architectures renowned for their efficacy in text similarity computations. Two primary models were employed for this endeavor.
For the Bi-Encoder model, we used \href{https://huggingface.co/sentence-transformers/all-MiniLM-L6-v2}{MiniLM model}, which was fine-tuned on an extensive set of 1 billion training pairs, this model excels in generating sentence embeddings suitable for our task.
For the Cross-Encoder model, we used \href{https://huggingface.co/cross-encoder/ms-marco-MiniLM-L-12-v2}{MiniLM model} provided from Sentence Transformers, which is trained on MS Marco passage ranking task. 

For bi-encoder and cross-encoder models, as they return similarity scores, we decide the threshold and determine whether atomic facts are present in the context of the resultant similarity score surpasses this threshold.
When deciding the threshold of the similarity score, we use the threshold that shows the highest correlation with humans. For the bi-encoder model, we use 0.4 (from a range of 0 to 1) as the threshold and for the cross-encoder model, we use 6 as the threshold. For both cases, we could see that the correlation tends to increase and decrease from a certain value, where the peak is the threshold value.

We further experiment over training cross-encoder MiniLM model with our dataset, pairs of input context, and atomic facts extracted from the context. However, due to the lack of diversity and a much smaller number of datasets compared to MS Marco, it showed lower human correlation (76.4), we used the released pretrained model as $M_{eval}$.

\section{Inference}
\label{app: inference}

\subsection{Model Details}
Llama2-chat is based on Llama2 and is optimized for dialogue using RLHF.
Vicuna\footnote{For 7B and 13B, we used version 1.5 and for 33B, we used version 1.3, where v1.5 is tuned on top of Llama2 and v1.3 is tuned on top of Llama1} is Llama2 finetuned on the outputs from ChatGPT available through ShareGPT. 
\textsc{T\"ulu1} and \textsc{T\"ulu2} are a Llama fine-tuned on mixture of human and machine-generated instructions and responses; \textsc{T\"ulu1} and \textsc{T\"ulu2} are finetuned on top of Llama1 and Llama2, respectively. Please note that \textsc{T\"ulu2} is finetuned on more larger dataset compared to \textsc{T\"ulu1}.
Falcon is trained on 1,000B tokens of RefinedWeb, and Falcon-Instruct is an instruction-tuned version of Falcon.
Mistral 
Models are selected to see the effect of instruction tuning, model size, and RLHF.

\subsection{Input Format} 
Figure~\ref{fig: input} shows the input format we used to generate all responses. Please note that for TULU, we changed the input format to match the format during training. ``<|user|> {instruction} <|assistant|>''
\begin{figure*}[t!]
{
\begin{tcolorbox}[width=0.99\textwidth, halign title=center, title = {Input Format for Evaluation}]
Generate an [answer] to the given [question] in full sentence by utilizing all necessary information in given [context] and limiting the utilized information to that [context]. Provide all information you utilize from given [context] to answer the question.\newline\newline
[context]\newline
Title: Greece [https://en.wikipedia.org/wiki/Greece]\newline
The country's rich historical legacy is reflected in part by its 18 UNESCO World Heritage Sites. Greece is a unitary parliamentary republic, and a developed country, with an advanced high-income economy. Its economy is the second largest in the Balkans, where it is an important regional investor. A founding member of the United Nations, Greece was the tenth member to join the European Communities (precursor to the European Union) and has been part of the Eurozone since 2001.\newline\newline
Title: Germany [https://en.wikipedia.org/wiki/Germany]\newline
After the fall of communist led-government in East Germany, German reunification saw the former East German states join the Federal Republic of Germany on 3 October 1990—becoming a federal parliamentary republic. Germany has been described as a great power with a strong economy; it has the largest economy in Europe, the world's fourth-largest economy by nominal GDP and the fifth-largest by PPP. As a global power in industrial, scientific and technological sectors, it is both the world's third-largest exporter and importer.\newline\newline
[question]\newline
Compare the economic rank of Germany and Greece.\newline\newline
Don't Forget that you have to generate an [answer] to the given [question] in full sentence by utilizing all necessary information in given [context] and information only from the [context]. Also, provide all information you utilize from given [context]\newline\newline
[answer]\newline
\end{tcolorbox}
}
\caption{Input format to generate response}
\label{fig: input}
\end{figure*}

\subsection{Inference Configuration}
In our research, we standardize the maximum input and output lengths at 2048 tokens for all experiments, except for those examining the effect of context length, where the maximum is extended to 4096 tokens. To ensure consistency across various model architectures, we apply 4-bit quantization during all experimental procedures.
We keep the generation configuration as same as the default configurations provided by Huggingface~\citep{Wolf2019HuggingFacesTS}. Specifically, for the Falcon, Llama2, and Vicuna models, we implement top-k sampling with a k value of 10. For the TULU model, we set the sampling temperature to 0.6.

\section{Results}

\subsection{Correlation between MMLU and Grounding Performance} \label{app: MMLU}
\begin{figure}[ht!]
    \centering
    \begin{minipage}[b]{0.49\textwidth}
    \includegraphics[width=\textwidth]{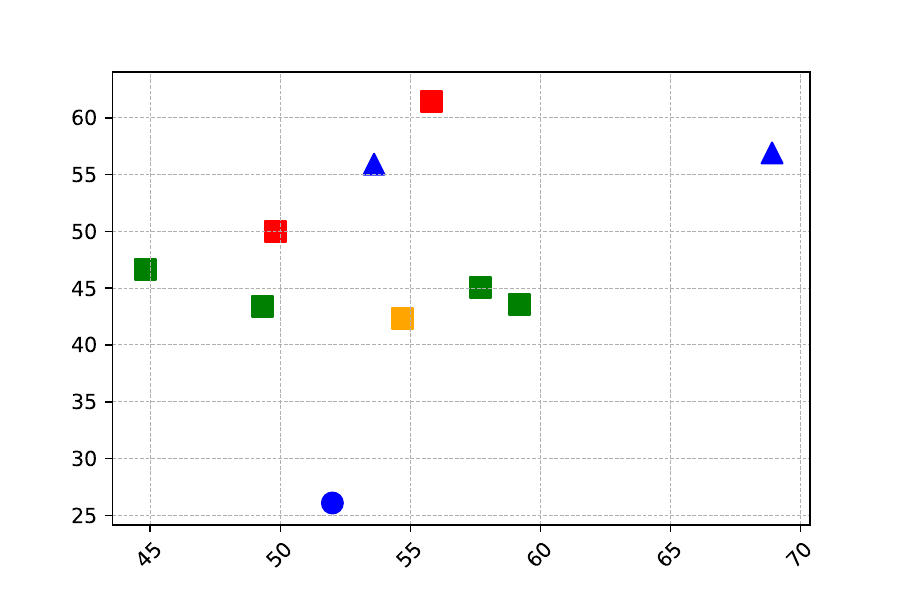}
    \caption{Correlation between MMLU performance and grounding performance: there is a weak correlation between the two.}
    \label{fig:app_MMLU}
    \end{minipage}
\end{figure}

To determine if grounding performance is strongly dependent on instruction-following ability, we see the correlation between grounding performance with performance on the MMLU benchmark~\citep{Hendrycks2020MeasuringMM}. MMLU is a widely used benchmark for the evaluation of instruction-tuned models~\citep{Sun2023EvaluatingTZ,Wang2023HowFC}, that requires a model to follow problem instructions over 57 subjects including STEM, humanities, social sciences, and more. The right figure in Figure~\ref{fig:app_MMLU} shows that there is a weak correlation between grounding abilities and MMLU scores\footnote{pearson correlation coefficient between grounding and MMLU performance is 0.32}. This suggests that grounding performance does not appear to be strongly reliant on the capacity to adhere to instructions.

\subsection{Grounding performance by different query and context characteristics}
\label{app: revised_gold}
\begin{figure}[ht!]
    \centering
    \begin{minipage}[b]{0.49\textwidth}
    \includegraphics[width=\textwidth]{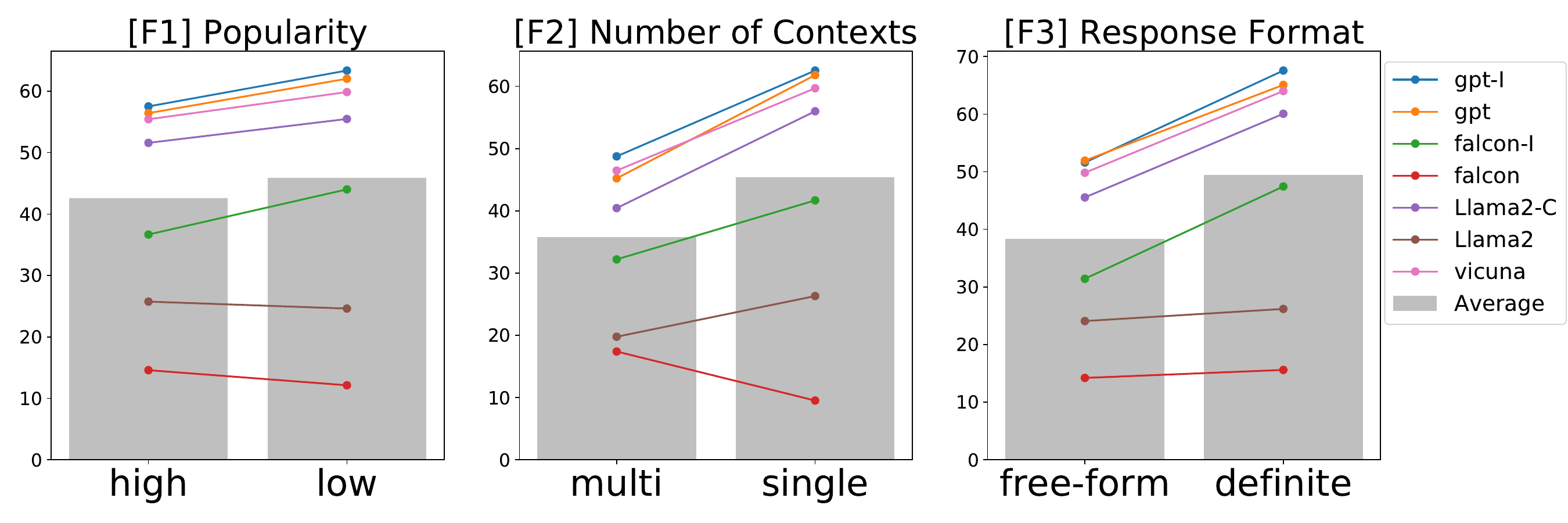}
    \caption{Details of Grounding performance by the characteristics of queries and contexts in \textit{Conflict-Gold}. \_I indicates instruction tuned version and \_C is those with RLHF tuned. Llama2 and vicuna is 13B, falcon is 40B model.}
    \label{fig:all_model.f1.revised}
    \end{minipage}
\end{figure}

Table~\ref{table: original_gold_specifics} shows the performance of models in \textit{Original-Gold}, Table~\ref{table: revised_gold_specifics} (Figure~\ref{fig:all_model.f1.revised}) shows the performance in \textit{Conflict-Gold}, Table~\ref{table: original_dist_specifics} shows the performance in \textit{Original-Dist}, and Table~\ref{table: revised_dist_specifics} shows the performance in \textit{Conflict-Dist}.
All dataset setting shows a similar trend with \textit{Original-Gold}. Vicuna-13b shows the highest performance over all open-sourced dataset. Grounding performance of pop high shows lower performance over pop low as models tend to utilize knowledge from given context more when it is not familiar with the knowledge~\citep{Mallen2022WhenNT}. Queries with single context (Single) show high grounding performance over queries that needs multiple context (Multi) since it is much easier and shorter; queries in Multi set often needs reasoning ability. 

\begin{table*}[ht!]
\centering
\fontsize{7.5}{10}\selectfont
    \begin{tabular}{ccc|cc|cc|cc}
    \toprule
         &  &  &  \multicolumn{2}{c}{$\mathcal{F}_1$} & \multicolumn{2}{c}{$\mathcal{F}_2$} & \multicolumn{2}{c}{$\mathcal{F}_3$} \\
    \midrule
        Model & Size & Grounding Perf. & High & Low & Free-Form & Definite & Single & Multi \\

    \midrule
    \multirow{3}{*}{Vicuna}
     & 7 & 50.01 & 45.31 & 55.39 & 39.99 & 58.5 & 51.94 & 40.4 \\
     & 33& 44.71 & 43.75 & 45.81 & 35.46 & 52.54 & 46.21 & 37.23 \\
     & 13  & 61.44 & 59.85 & 63.25 & 52.55 & 68.96 & 64.07 & 48.27 \\
     \midrule
    \multirow{4}{*}{\textsc{T\"ulu1}}
    & 7 & 46.67 & 46.32 & 47.08 & 50.84 & 43.15 & 49.1 & 34.54 \\
     & 13  & 43.42 & 41.15 & 46.01 & 51.63 & 36.46 & 46.03 & 30.35 \\
    & 30  & 45.06 & 45.19 & 44.92 & 52.12 & 39.09 & 46.86 & 36.07 \\
     & 65  & 43.58 & 41.58 & 45.86 & 53.11 & 35.52 & 46.04 & 31.27 \\
      \midrule
    \multirow{3}{*}{\textsc{T\"ulu2}}
     & 7  & 58.57 & 56.22 & 61.24 & 49.09 & 66.58 & 60.99 & 46.46 \\
    & 70  & 59.61 & 57.09 & 62.48 & 53.27 & 64.97 & 62.77 & 43.8 \\
    & 13  & 62.29 & 59.97 & 64.95 & 55.58 & 67.98 & 65.6 & 45.77 \\
     \midrule
    \multirow{3}{*}{\textsc{T\"ulu2}-D}
     & 7  & 51.46 & 48.24 & 55.15 & 41.14 & 60.2 & 53.01 & 43.75 \\
     & 70 & 58.02 & 57.03 & 59.14 & 50.2 & 64.63 & 60.55 & 45.36 \\
     & 13  & 60.11 & 57.32 & 63.29 & 50.11 & 68.57 & 62.76 & 46.86 \\    
      \midrule
    Mistral-I & 7  & 60.26 & 57.82 & 63.04 & 53.97 & 65.57 & 63.05 & 46.29 \\
    Zephyr & 7  & 54.72 & 52.57 & 57.18 & 42.86 & 64.75 & 56.89 & 43.89 \\
     \midrule
    \multirow{3}{*}{Llama2-C}
     & 7 & 51.63 & 47.81 & 56 & 38.26 & 62.95 & 53.97 & 39.93 \\
    & 13 & 55.91 & 54.57 & 57.44 & 45.58 & 64.65 & 58.38 & 43.54 \\
    & 70  & 56.9 & 56.53 & 57.32 & 50.53 & 62.29 & 58.62 & 48.32 \\
     \midrule
    Llama2 & 13  & 26.09 & 23.21 & 29.38 & 23.04 & 28.67 & 28.05 & 16.31 \\
     \midrule
    GPT & - & 61.01 & 60.06 & 62.11 & 52.94 & 67.85 & 63.68 & 47.68 \\
    GPT-I & - & 65.69 & 63.23 & 68.5 & 56.92 & 73.11 & 68.36 & 52.31 \\
     \midrule
     \multirow{2}{*}{Falcon}
    & 40 & 18.92 & 18.16 & 19.8 & 19.86 & 18.13 & 19.64 & 15.34 \\
     & 180 & 26.4 & 28.38 & 24.14 & 23.70 & 28.69 & 26.88 & 24.01 \\
         \midrule
     \multirow{2}{*}{Falcon-I}
     & 40 & 42.35 & 38.36 & 46.91 & 33.15 & 50.13 & 44.61 & 31.03 \\
      & 180 & 46.16 & 43.54 & 49.14 & 40.52 & 50.92 & 48.74 & 33.23 \\
    \bottomrule
    \end{tabular}
\caption
     {\fontsize{6.5}{10}\footnotesize Specific performance of \textit{Original-Gold}. Best from all models in \textbf{Bold} and best from open-sourced models in \underline{underline}.}
\label{table: original_gold_specifics}
\end{table*}

\begin{table*}[ht!]
\centering
\fontsize{7.5}{10}\selectfont
    \begin{tabular}{ccc|cc|cc|cc}
    \toprule
         &  &  &  \multicolumn{2}{c}{$\mathcal{F}_1$} & \multicolumn{2}{c}{$\mathcal{F}_2$} & \multicolumn{2}{c}{$\mathcal{F}_3$} \\
    \midrule
        Model & Size & Grounding Perf. & High & Low & Free-Form & Definite & Single & Multi \\

    \midrule
    \multirow{3}{*}{Vicuna}
     & 7 & 47.98 & 46.08 & 50.14 & 38.67 & 55.85 & 50.05 & 37.6 \\
     & 13 & 57.5 & 55.43 & 59.86 & 49.82 & 64 & 59.7 & 46.47 \\
     & 33 & 40.32 & 38.84 & 42.02 & 40 & 40.6 & 41.36 & 35.13 \\
        \midrule
    \multirow{3}{*}{\textsc{T\"ulu1}}
    & 7 & 46.52 & 46.75 & 46.26 & 48.27 & 45.04 & 48.05 & 38.87 \\
    & 13 & 41.35 & 39.78 & 43.14 & 45.95 & 37.46 & 43.68 & 29.71 \\
    & 30 & 43.95 & 45 & 42.75 & 49.29 & 39.43 & 45.51 & 36.14 \\
    & 65 & 39.47 & 39.78 & 39.12 & 50.59 & 30.07 & 40.77 & 32.97 \\
        \midrule
    \multirow{3}{*}{\textsc{T\"ulu2}}
     & 7 & 54.86 & 52.22 & 57.88 & 47.41 & 61.16 & 57.4 & 42.19 \\
     & 13 & 61.9 & 59.7 & 64.42 & 57.02 & 66.03 & 64.35 & 49.67 \\  
     & 70 & 59.93 & 57.87 & 62.29 & 53.64 & 65.26 & 61.15 & 53.83 \\
    \midrule
    \multirow{3}{*}{\textsc{T\"ulu2}-D}
    & 7 & 51.36 & 48.66 & 54.43 & 40.28 & 60.73 & 52.73 & 44.46 \\
    & 13 & 58.03 & 55.82 & 60.55 & 48.34 & 66.22 & 60.03 & 48.01 \\
    & 70 & 58.07 & 56.35 & 60.04 & 49.33 & 65.47 & 59.88 & 49.04 \\
    \midrule
    Mistral-I & 7 & 59.83 & 57.32 & 62.69 & 54.39 & 64.43 & 61.92 & 49.38 \\
    Zephyr & 7 & 52.37 & 50.34 & 54.69 & 44.03 & 59.42 & 54.36 & 42.4 \\
    \midrule
    \multirow{3}{*}{Llama2-C}
     & 7 & 45.95 & 42.79 & 49.58 & 35.2 & 55.05 & 47.68 & 37.35 \\
    & 13 & 53.41 & 51.59 & 55.48 & 45.54 & 60.06 & 56 & 40.44 \\
    & 70 \\
    \midrule
    Llama2 & 13 & 25.22 & 25.75 & 24.62 & 24.08 & 26.19 & 26.31 & 19.77 \\
    \midrule
    GPT & - & 59.04 & 56.43 & 62.03 & 51.93 & 65.07 & 61.81 & 45.22 \\
    GPT-I & - & 60.25 & 57.52 & 63.36 & 51.6 & 67.56 & 62.54 & 48.75 \\
    \midrule
    \multirow{2}{*}{Falcon}
     & 40 & 23.63 & 22.13 & 25.34 & 24.37 & 23 & 24.47 & 19.42 \\
      & 180 & 25.59 & 25.52 & 25.67 & 23.34 & 27.5 & 27.33 & 16.92 \\
        \midrule
    \multirow{2}{*}{Falcon-I}
     & 40 & 40.1 & 36.67 & 44.02 & 31.42 & 47.44 & 41.68 & 32.2 \\
       & 180 & 45.31 & 41.97 & 49.12 & 37.35 & 50.2 & 46.19 & 34.9 \\
    \bottomrule
    \end{tabular}
\caption
     {\fontsize{6.5}{10}\footnotesize Specific performance of \textit{Conflict-Gold}. Best from all models in \textbf{Bold} and best from open-sourced models in \underline{underline}.}
\label{table: revised_gold_specifics}
\end{table*}

\begin{table*}[ht!]
\centering
\fontsize{7.5}{10}\selectfont
    \begin{tabular}{ccc|cc|cc|cc}
    \toprule
         &  &  &  \multicolumn{2}{c}{$\mathcal{F}_1$} & \multicolumn{2}{c}{$\mathcal{F}_2$} & \multicolumn{2}{c}{$\mathcal{F}_3$} \\
    \midrule
        Model & Size & Grounding Perf. & High & Low & Free-Form & Definite & Single & Multi \\

    \midrule
    \multirow{2}{*}{Vicuna}
     & 7 & 45.01 & 40.24 & 50.45 & 38.58 & 50.44 & 47.29 & 33.6 \\
     & 13 & 57.46 & 55.91 & 59.23 & 49.13 & 64.51 & 59.12 & 49.17 \\
     \midrule
     \multirow{3}{*}{\textsc{T\"ulu1}}
     & 7 & 44.57 & 40.84 & 48.82 & 44.88 & 44.3 & 47.61 & 29.36 \\
     & 13 & 41.95 & 38.41 & 46 & 45.24 & 39.17 & 44.9 & 27.21 \\
      & 30 & 40.95 & 40.77 & 41.16 & 49.56 & 33.67 & 43.18 & 29.81 \\
     & 65 & 39.12 & 40.26 & 37.82 & 48.68 & 31.03 & 41.04 & 29.5 \\
     \midrule
     \multirow{3}{*}{\textsc{T\"ulu2}}
     & 7 & 54.9 & 52.66 & 57.46 & 47.18 & 61.43 & 57.69 & 40.94 \\
     & 13 & 55.27 & 52.66 & 58.26 & 52.04 & 58 & 58.12 & 41.05 \\
    & 70 & 53.43 & 53.3 & 53.58 & 52.96 & 53.83 & 56.46 & 38.26 \\
    \midrule
     \multirow{3}{*}{\textsc{T\"ulu2}-D}
     & 7 & 45.26 & 42.96 & 47.9 & 36.86 & 52.37 & 46.7 & 38.06 \\
    & 13 & 53.98 & 52.03 & 56.2 & 45.57 & 61.08 & 56.18 & 42.94 \\
     & 70 & 55.41 & 53.61 & 57.47 & 47.9 & 61.76 & 58.24 & 41.27 \\
     \midrule
    Mistral-I & 7 & 54.87 & 53.07 & 56.92 & 49.32 & 59.56 & 58.37 & 37.36 \\
        Zephyr & 7 & 53.66 & 50.56 & 57.21 & 44.29 & 61.58 & 56.52 & 39.35 \\
    \midrule
        \multirow{3}{*}{Llama2-C}
     & 7 & 45.14 & 43.9 & 46.55 & 37.14 & 51.91 & 47.57 & 32.98 \\
    & 70 & 56.24 & 54.17 & 58.61 & 47.9 & 63.3 & 58.89 & 43.01 \\
     & 13 & 35.83 & 35.5 & 36.21 & 35.83 & 35.84 & 37.23 & 28.88 \\
     \midrule
    Llama2 & 13 & 21.68 & 21.55 & 21.83 & 19.71 & 23.35 & 22.53 & 17.44 \\
    \midrule
    GPT & 0 & 56.78 & 54.25 & 59.66 & 47.77 & 64.4 & 59.99 & 40.72 \\
    GPT-I & 0 & 56.87 & 55.67 & 58.24 & 47.2 & 65.05 & 59.96 & 41.41 \\
    \midrule
    Falcon-I & 40 & 36.33 & 33.21 & 39.9 & 29.88 & 41.79 & 38.18 & 27.07 \\
    \bottomrule
    \end{tabular}
\caption
     {\fontsize{6.5}{10}\footnotesize Specific performance of \textit{Original-Dist}. Best from all models in \textbf{Bold} and best from open-sourced models in \underline{underline}.}
\label{table: original_dist_specifics}
\end{table*}

\begin{table*}[ht!]
\centering
\fontsize{7.5}{10}\selectfont
    \begin{tabular}{ccc|cc|cc|cc}
    \toprule
         &  &  &  \multicolumn{2}{c}{$\mathcal{F}_1$} & \multicolumn{2}{c}{$\mathcal{F}_2$} & \multicolumn{2}{c}{$\mathcal{F}_3$} \\
    \midrule
        Model & Size & Grounding Perf. & High & Low & Free-Form & Definite & Single & Multi \\

    \multirow{2}{*}{Vicuna}
     & 7 & 39.76 & 39.18 & 40.42 & 33.39 & 45.15 & 41.53 & 30.9  \\
    & 13 & 55.04 & 52.76 & 57.65 & 46.8 & 62.02 & 58.48 & 37.88  \\
     \multirow{3}{*}{\textsc{T\"ulu1}}
    & 7 & 44.39 & 41.2 & 48.04 & 45.51 & 43.44 & 46.89 & 31.92  \\
     & 13 & 40.37 & 39.03 & 41.9 & 45.77 & 35.81 & 43.04 & 27.02  \\
     & 65 & 36.3 & 36.96 & 35.55 & 48.76 & 25.75 & 38.33 & 26.14  \\
     & 30 & 40.87 & 39.78 & 42.1 & 47.04 & 35.64 & 42.61 & 32.14  \\
      \multirow{3}{*}{\textsc{T\"ulu2}-D}
     & 7 & 41.43 & 39.63 & 43.47 & 33.29 & 48.31 & 42.27 & 37.19 \\
     & 70 & 55.06 & 53.88 & 56.42 & 47.51 & 61.45 & 57.34 & 43.70  \\
     & 13 & 54.19 & 52.11 & 56.56 & 45.41 & 61.62 & 56.48 & 42.71  \\
       \multirow{3}{*}{\textsc{T\"ulu2}}
    & 7 & 47.92 & 45.12 & 51.13 & 42.4 & 52.6 & 50.41 & 35.47   \\
     & 70 & 52.38 & 49.72 & 55.41 & 50.87 & 53.65 & 54.48 & 41.86  \\
     & 13 & 50.41 & 47.13 & 54.16 & 48.66 & 51.9 & 52.44 & 40.27  \\
     \midrule
    Mistral-I & 7 & 54.28 & 51.51 & 57.44 & 47.83 & 59.73 & 57.23 & 39.49  \\
    Zephyr & 7 & 52.4 & 50.3 & 54.8 & 43.99 & 59.52 & 54.36 & 42.62 \\
    \midrule
       \multirow{3}{*}{Llama2-C}
     & 7 & 40.39 & 38.77 & 42.24 & 31.15 & 48.21 & 41.86 & 33.06  \\
     & 13 & 46.45 & 45.09 & 48 & 40.95 & 51.1 & 48.52 & 36.09  \\
     & 70 & 54.36 & 53.43 & 55.42 & 47.7 & 60 & 56.63 & 42.99  \\
     \midrule
    Llama2 & 13 & 19.3 & 19.17 & 19.44 & 20.38 & 18.38 & 20.03 & 15.64  \\
    \midrule
    GPT & - & 56.08 & 52.4 & 60.28 & 50.08 & 61.15 & 58.64 & 43.28  \\
    GPT-I & - & 54.54 & 53.61 & 55.6 & 48.53 & 59.62 & 56.56 & 44.41 \\
    \midrule
    Falcon & 40 & 12.14 & 10.27 & 14.27 & 14.52 & 10.13 & 12.56 & 10.02 \\
    Falcon-I & 40 & 32.6 & 28.6 & 37.16 & 27.69 & 36.75 & 34.47 & 23.21 \\
    \bottomrule
    \end{tabular}
\caption
     {\fontsize{6.5}{10}\footnotesize Specific performance of \textit{Conflict-Dist}. Best from all models in \textbf{bold} and best from open-sourced models in \underline{underline}.}
\label{table: revised_dist_specifics}
\end{table*}

\subsection{Precision and Recall} \label{app: prec_and_recall}
\begin{figure}[t!]
    \centering
    \begin{minipage}[b]{0.45\textwidth}
    \includegraphics[width=\textwidth]{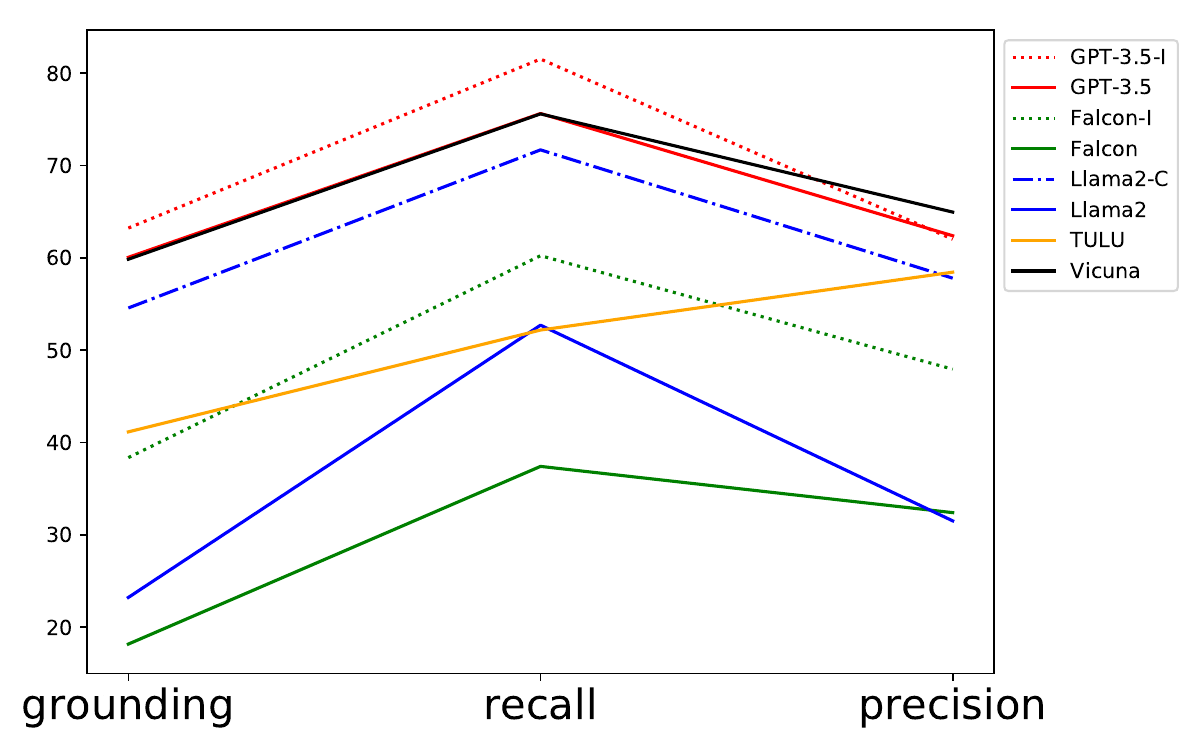}
    \caption{Performance of grounding performance, precision, and recall in \textit{Original-Gold}}
    \label{fig:all_model.prec_recall.original}
    \end{minipage}
    \begin{minipage}[b]{0.45\textwidth}
    \includegraphics[width=\textwidth]{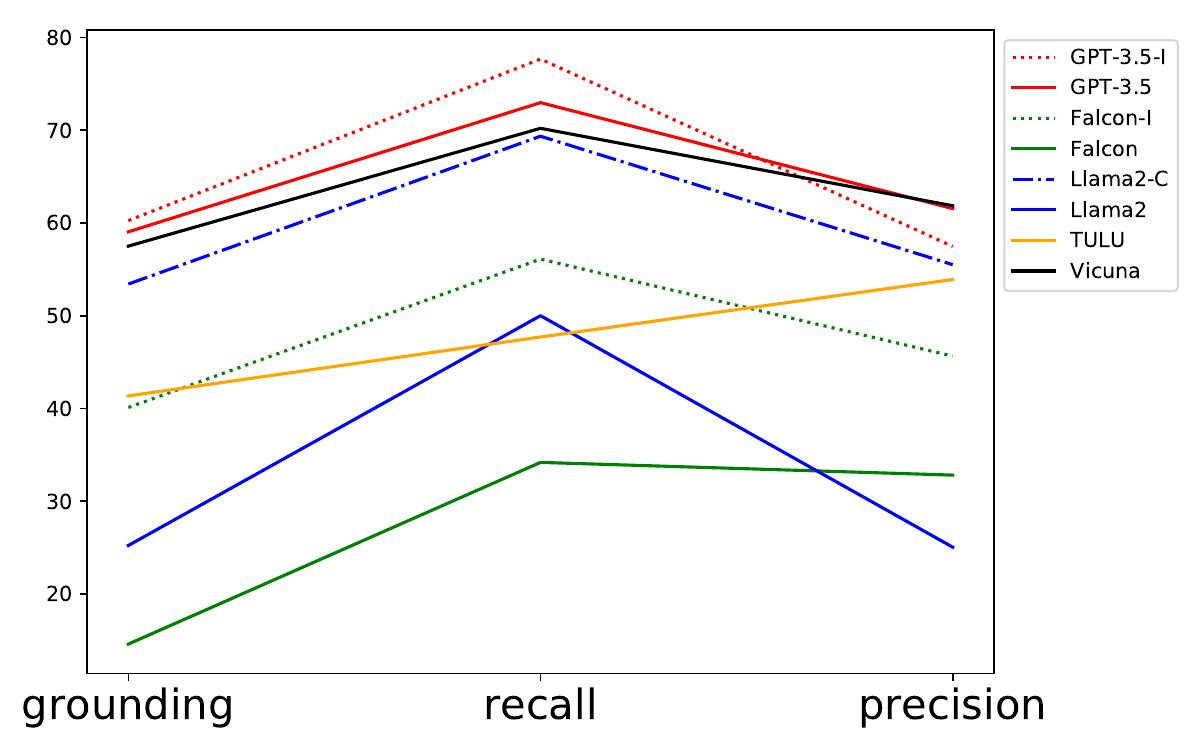}
    \caption{Performance of grounding performance, precision, and recall in \textit{Conflict-Gold}}
    \label{fig:all_model.prec_recall.revised}
    \end{minipage}
\end{figure}
Figure~\ref{fig:all_model.prec_recall.original} presents the precision and recall metrics for the \textit{Original-Gold} dataset, whereas Figure~\ref{fig:all_model.prec_recall.revised} displays the same for the \textit{Conflict-Gold} dataset. Precision is measured to determine if the source of atomic facts in the knowledge base is the input context rather than external sources. Recall, on the other hand, assesses whether all essential knowledge (gold atomic facts) is included in the generated response. From the results for both datasets, it is evident that recall outperforms precision, suggesting that the model tends to incorporate knowledge beyond the provided information when evaluating them in a fine-grained manner.

\subsection{Larger models Tend to Show Higher Degradation with Distractor Contexts} 
\label{app: tulu_distractor}
\begin{figure}[t!]
    \centering
    \begin{minipage}[b]{0.45\textwidth}
    \includegraphics[width=\textwidth]{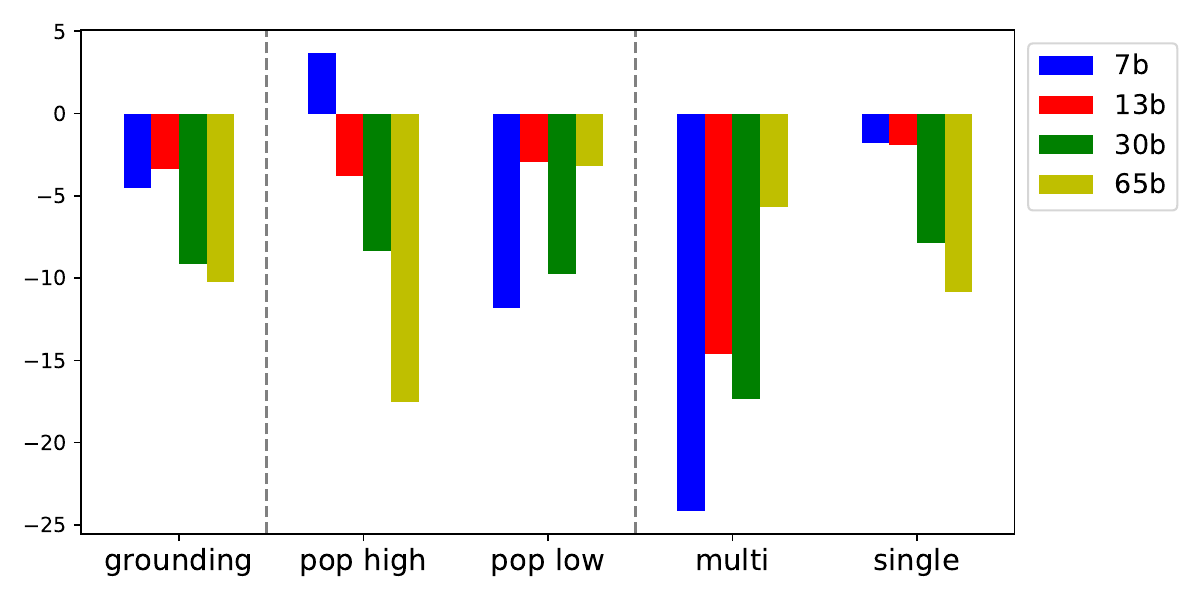}
    \caption{Reduction rate in grounding performance when adding distractor contexts}
    \label{fig:tulu.ori_ori_dist}
    \end{minipage}
\vspace{-1em}
\end{figure}

Figure~\ref{fig:tulu.ori_ori_dist} demonstrates that larger models tend to show higher degradation when distractor contexts are added. 
The most significant reduction is observed in recall rather than precision (Appendix~\ref{app: prec_and_recall}), suggesting that the models often default to providing only the answer without detailed explanations. The lower grounding performance for these queries is largely due to this tendency to omit specific details.
Conversely, for queries requiring multiple contexts (multi), a different pattern emerges: smaller models exhibit more significant performance drops. These multi-context queries are inherently more complex, often necessitating advanced reasoning or a deeper understanding of the overall context, leading to a steeper decline in grounding performance for smaller models as the task difficulty increases.

\subsection{Average Number of Contexts for Distractor Settings} \label{app: distractor}
In our datasets, \textit{Original-Gold} and \textit{Conflict-Gold}, the contexts exhibit an average token length of 335, which is comparatively brief. To address this, we incorporate distractor contexts into our analysis. These distractors are contextually relevant to the queries but do not contain the gold atomic facts. As illustrated in Figure~\ref{fig:dist_length}, the average number of contexts per query is 3.3, 11.1, 19.1, and 24.0. These values correspond to the circle markers shown in the figure, indicating a varied context distribution in our dataset.

\subsection{Performance on Answer Accuracy} \label{app: answer_acc}

\begin{table*}[t!]
\centering
\fontsize{7.5}{10}\selectfont
    \begin{tabular}{cccccccccccccc}
    \toprule
        Size & \multicolumn{2}{c}{7B} & \multicolumn{4}{c}{13B} & 30B  & \multicolumn{2}{c}{40B} & 65B & \multicolumn{2}{c}{UNK}  \\
            \cmidrule(lr){2-3} \cmidrule(lr){4-7} \cmidrule(lr){9-10} \cmidrule(lr){12-13}
         $M_{pred}$  & Vicuna & TULU & Llama2 & Llama2-chat & Vicuna & TULU & TULU & Falcon & Falcon-I & TULU & GPT-3.5 & GPT-3.5-I\\
    \midrule
    Without Contexts & 16.40 & 14.81 & 28.91 & \underline{35.98} & 30.40 & 15.67 & 28.90 & 33.91 & 31.85 & 22.49 & \textbf{47.11} & 45.55 \\
    Original-Gold & 83.06 & 77.83 & 81.56 & 84.79 & \underline{86.57} & 82.62 & 83.74 & 70.19 & 82.38 & 83.38 & 88.16 & \textbf{91.31} \\
    Original-Dist & 70.88 & 70.83 & 72.85 & 80.26 & \underline{81.50} & 77.27 & 77.33 & 63.2 & 70.26 & 79.51 & 87.00 & \textbf{88.01} \\
    Conflict-Gold & 76.19 & 76.94 & 77.26 & \underline{81.36} & 80.90 & 76.64 & 76.82 & 58.84 & 71.49 & 78.29 & \textbf{86.13} & 84.79 \\
    Conflict-Dist & 66.91 & 64.67 & 57.88 & 55.51 & \underline{73.49} & 69.91 & 71.75 & 55.51 & 60.10 & 70.97 & 79.95 & \textbf{83.32} \\
    \bottomrule
    \end{tabular}
\caption
     {\fontsize{6.5}{10}\footnotesize Answer Accuracy of twelve different models. For each setting, the best in \textbf{bold} and the best of open-sourced models in \underline{underline}.}
\label{table: ans_acc}
\end{table*}

Table~\ref{table: ans_acc} shows the answer accuracy of models across five settings. 
Diving into performance based on input context and question traits reveals key patterns. Without external contexts, high-popularity questions achieve a 32.6\% accuracy, outpacing low-popularity ones at 26.8\%. However, this changes with gold contexts: low-popularity questions slightly edge out at 83.4\% over the 83.2\% for high-popularity ones. This likely stems from models leaning more on given contexts when unsure, mirroring \citet{Mallen2022WhenNT} findings.
Regarding the number of input contexts, queries requiring multiple contexts generally fare worse than those with one. The gap is wider for smaller models (under 40b parameters): they experience a 23.7\% drop, while larger models see only a 13.1\% dip. This underscores bigger models' superior multi-context comprehension and reasoning capacity.
We believe this discrepancy highlights a larger model's enhanced reasoning capacity and its ability to better understand multiple contexts.
Lastly, revising or adding distractors to contexts affects accuracy. It declines notably with both actions, with a steeper 12.4\% fall when distractors are added to modified contexts, compared to 7.8\% for original contexts.

\begin{figure*}[t!]
{
\begin{tcolorbox}[width=0.99\textwidth, halign title=center, title = {Instructions for evaluation of fluency}]
You will be given one response written for a instruction.\newline
\newline
Your task is to rate the response on one metric.\newline
\newline
Please make sure you read and understand these instructions carefully. Please keep this document open while reviewing, and refer to it as needed.\newline
\newline
\newline
Evaluation Criteria:\newline
\newline
Fluency (1-5): the quality of the response upon the Input in terms of grammar, spelling, punctuation, word choice, and sentence structure. The response should not contain any unnatural symbols. \newline
\newline
- 1: Very Poor.     The response is mostly incoherent with severe issues in grammar, spelling, punctuation, word choice, sentence structure, and contains unnatural symbols.\newline
- 2: Below Average. The response is understandable with effort; numerous errors in grammar, spelling, punctuation, word choice, and sentence structure; may have unnatural symbols.\newline
- 3: Average.       The response is understandable with occasional errors in grammar, spelling, punctuation, word choice, or sentence structure; no unnatural symbols.\newline
- 4: Above Average. The response is mostly fluent with very few errors; clear and easy to understand; no unnatural symbols.\newline
- 5: Excellent.     The response is perfectly fluent; free from any errors; clear, concise, and natural with no unnatural symbols.\newline
\newline
Evaluation Steps:\newline
1. Read the given response thoroughly.\newline
2. Check for any spelling mistakes.\newline
3. Examine the grammar and sentence structure. Look for incorrect verb conjugations, misplaced modifiers, and other grammatical mistakes.\newline
4. Ensure that punctuation is used correctly. Check for missing or misused commas, periods, semicolons, etc.\newline
5. Evaluate the word choice. Are the words appropriate for the context? Are there any words that sound unnatural or out of place?\newline
6. Confirm that there are no unnatural symbols or characters in the response.\newline
7. Based on the observations, rate the fluency of the response using the provided scale (1-5).\newline
\newline
\newline
Example:\newline
\newline
\newline
Response:\newline
\newline
\{response\}\newline
\newline
\newline
Evaluation Form (scores ONLY):\newline
\newline
Fluency (1-5):\newline
\end{tcolorbox}
}

\caption{Instructions for Evaluation of Fluency} 
\label{fig:fluency_instruction} 
\end{figure*}
\subsection{Performance on Fluency}
\label{app: fluency}
\begin{table}[t!]
\centering
\fontsize{7.5}{10}\selectfont
    \begin{tabular}{cccccc}
    \toprule
         \multicolumn{4}{c}{13B} & 30B  & 40B \\
            \cmidrule(lr){1-4} 
           Llama & Llama-C & Vicuna & TULU & TULU & Falcon-I \\
    \midrule
     3.66 & 4.96 & 4.94 & 4.87 & 4.92 & 4.97 \\
    \bottomrule
    \end{tabular}
\caption
     {\fontsize{6.5}{10}\footnotesize Fluency of LLMs measured by G-EVAL. Here, Llama is Llama2 and Llama-C is Llama2-Chat and Falcon-I is Falcon-Instruct.}
\label{table: fluency}
\end{table}
Our grounding assessment risks being skewed by responses that merely extract and piece together fragments of external knowledge. To counter this, we evaluate the fluency of the generated responses to determine whether they are formulated in a naturally coherent manner.
We employ G-EVAL~\citep{Liu2023GEvalNE} to evaluate fluency, a framework that uses large language models in a chain-of-thought and form-filling paradigm. This fluency metric is particularly applied to queries requiring free-form answers as we observed that some models tend to produce only direct answers thus difficult to evaluate the fluency. 
Table~\ref{table: fluency} shows the fluency scores of six LLMs.
Notably, all models demonstrate high fluency, with Llama2 exhibiting the lowest score. This is attributed to its lack of instruction tuning, leading it to generate longer, less relevant sentences reminiscent of its pretraining data. The instructions used to evaluate fluency are detailed in Figure~\ref{fig:fluency_instruction}.

\end{document}